\documentclass{article}
\usepackage[utf8]{inputenc}
\usepackage[letterpaper]{geometry}
\usepackage{libertine}
\usepackage{libertinust1math}
\usepackage[T1]{fontenc}
\usepackage{amsmath}
\usepackage{amssymb,amsthm}
\usepackage{xcolor}
\usepackage[ruled,vlined,linesnumbered]{algorithm2e}
\usepackage[caption=false,font=footnotesize]{subfig}
\usepackage[]{hyperref}
\usepackage{graphicx}
\usepackage[square,authoryear]{natbib}
\usepackage[]{authblk}
\usepackage{url}
\usepackage{amsmath}
\usepackage{amssymb,amsthm}
\usepackage{xcolor}
\usepackage[ruled,vlined,linesnumbered]{algorithm2e}

\SetKwBlock{KwInitialize}{Initialize:}{}
\usepackage[caption=false,font=footnotesize]{subfig}
\usepackage{booktabs}

\newtheorem{theorem}{Theorem}

\newtheorem{lemma}[theorem]{Lemma}
\newtheorem{definition}{Definition}
\newtheorem{assumption}{Assumption}

\usepackage{etoolbox}

\makeatletter
\patchcmd{\@algocf@start}
  {-1.5em}
  {10pt}
  {}{}
\makeatother

\SetAlgorithmName{Procedure}{procedure}{List of Procedures}

\title{Bandit Data-Driven Optimization\thanks{This is the complete version of the paper. A version of this paper is published at AAAI-22.}}
\author[1,2]{Zheyuan Ryan Shi}
\author[1]{Zhiwei Steven Wu}
\author[1]{Rayid Ghani}
\author[1]{Fei Fang}
\affil[1]{ Carnegie Mellon University }
\affil[2]{98Connect}
\affil[ ]{\normalsize \{ryanshi, zstevenwu, rayid\}@cmu.edu, feif@cs.cmu.edu}

\date{}
\begin{document}

\maketitle

\begin{abstract}
Applications of machine learning in the non-profit and public sectors often feature an iterative workflow of data acquisition, prediction, and optimization of interventions.
There are four major pain points that a machine learning pipeline must overcome in order to be actually useful in these settings: small data, data collected only under the default intervention, unmodeled objectives due to communication gap, and unforeseen consequences of the intervention. 
In this paper, we introduce bandit data-driven optimization, the first iterative prediction-prescription framework to address these pain points.
Bandit data-driven optimization combines the advantages of online bandit learning and offline predictive analytics in an integrated framework. We propose PROOF, a novel algorithm for this framework and formally prove that it has no-regret. 
Using numerical simulations, we show that PROOF achieves superior performance than existing baseline. We also apply PROOF in a detailed case study of food rescue volunteer recommendation, and show that PROOF as a framework works well with the intricacies of ML models in real-world AI for non-profit and public sector applications.
\end{abstract}

\section{Introduction}
\label{sec:intro}

The success of modern machine learning (ML) largely lies in supervised learning, where one predicts some label $c$ given input feature $x$. 
Off-the-shelf predictive models have made their ways into numerous commercial applications. 
Deep learning has repeatedly advanced our ability to tell a cat from a dog. 
Such tangible progress has motivated the community to address more real-world societal challenges, and in particular, in the non-profit and public sectors.

\begin{figure*}[t]
    \centering
    \includegraphics[width=\columnwidth]{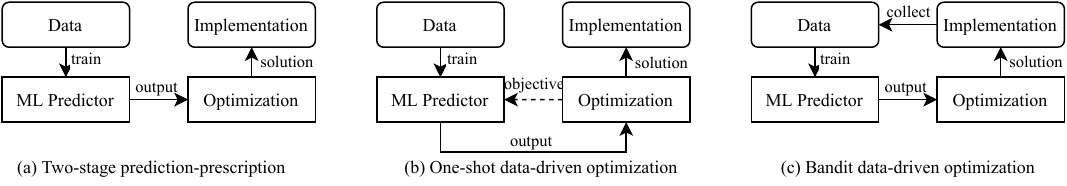}
    \caption{Paradigms of how ML systems are used in realistic settings.}
    \label{fig:paradigm}
\end{figure*}

Unfortunately, the success of ML often does not translate directly into a satisfactory solution to a real-world societal problem. One obvious reason is supervised learning focuses on prediction, yet real-world problems, by and large, need prescription. For example, rather than predict which households' water pipes are contaminated (labels) using construction data (features), municipal officials need to schedule inspections (interventions)~\citep{abernethy2018activeremediation}. The common practice is a two-stage procedure, as shown in Figure~\ref{fig:paradigm}a. After training an ML model, one makes prescriptive decisions based on some optimization problem parametrized by the prediction output.
In an emerging line of work on (one-shot) data-driven optimization, the learning problem is made aware of the downstream optimization objective through its loss function, gluing the two stages together~\citep{bertsimas2020predictive,elmachtoub2017smart}. We illustrate this in Figure~\ref{fig:paradigm}b.

However, this is still far from a complete picture. Figure~\ref{fig:paradigm}c shows a typical workflow in many AI for non-profit and public sector projects. After getting existing data which are often under a default intervention, a data scientist trains an ML model and then, based on it, recommends an intervention. Using the new data collected under the new intervention, the researcher updates the ML model and recommends a new intervention, so on and so forth, leading to an iterative process. The principles of these steps are often not aligned. Without a rigorous, integrated framework to guide the procedure, this could lead to operation inefficiency, missed expectations, dampened initiatives, and new barriers of mistrust which are not meant to be. 

While the reader might think this is workflow is universal across applications of ML, whether for profit or not, such an iterative process is especially prominent and necessary in the non-profit and public sectors due to the following key features of those applications distilled from existing research~\citep{perrault2019ai}.
First, there may not be enough data to begin with. Many of these domains do not have the luxury of millions of training examples. A small dataset at the beginning leads to inaccurate predictions and hence suboptimal decisions, but they will improve as we collect more data, as seen in, e.g., predicting poaching threats from patrol data and designing ranger patrols~\citep{gholami2019don}. Second, too often the initial dataset has some default intervention embedded, while the project's goal is to find the optimal intervention. For example,~\citet{shi2020improving} design a smartphone notification scheme for a volunteer-based platform but existing data are all collected under a default suboptimal scheme. If one expects the data distribution to vary across interventions, one has to try out some interventions and collect data under them.
Third, we may not perfectly know the the correct objective function to optimize. This is especially true considering the knowledge and communication gap between the data scientists and the domain practitioners. Fourth, the proposed interventions may have unexpected consequences. This hints at the inherent impossibility of fully modeling the problem in one shot. 

We propose the first iterative prediction-prescription framework, which we term as \emph{bandit data-driven optimization}. This framework combines the relative advantages of both online bandit learning and offline predictive analytics. We achieve this with our algorithm PRedict-then-Optimize with Optimism in Face of uncertainty (PROOF). PROOF is a modular algorithm which can work with a variety of predictive models and optimization problems. Under specific settings, we formally analyze its performance and show that PROOF achieves no-regret. In addition, we propose a variant of PROOF which handles the scenario where the intervention affects the data distribution and prove that it also enjoys no-regret. Using numerical simulations, we show that PROOF achieves much better performance than a pure bandit baseline. We also apply PROOF in a case study of a real-world AI for nonprofit project on food rescue volunteer engagement.

\section{Related Work}
\label{sec:relatedwork}

We propose bandit data-driven optimization to address the challenges we encountered in our previous work on AI for the non-profit and public sectors, because we found surprisingly no existing work that rigorously studies the iterative prediction-prescription procedure.
We explain below how several lines of work with similar goals fail to address the challenges we face, and summarize them in Table~\ref{tab:relatedwork}.

First, (one-shot) data-driven optimization aims to find the action $w^*$ that maximizes the expected value of objective function $p(c, w)$ given some feature $x$ where $c$ is a function of $x$, i.e. $w^* = \arg\max_w \mathbb E_{c|x} [p(c, w)]$ \citep{bertsimas2020predictive,ban2019big}. 
A popular approach is referred to as the predict-then-optimize framework~\citep{elmachtoub2017smart,kao2009directed}. There, one learns an ML predictor $f$ from data and then optimize $p(c,w)$ with the predicted label $c = f(x)$.
This entire literature assumes that the optimization objective is known a priori, which is often too good to be true. It also does not consider sequential settings and hence cannot adapt to new data. 
Meanwhile, inverse optimization~\citep{esfahani2018data,dong2018generalized} also does not apply to our problem, for the actions taken obviously have no definite relationship with the optimal action.
Our work touches on optimization under uncertainty~\citep{zheng2018active,chen2017nearly,balkanski2016power}. They learn the parameters of an optimization problem. However, they do not learn the data distribution or use the feature/label dataset that is so common in real-world applications like food rescue.

Contextual bandit is a proper setting for sequential decision making~\citep{lai1985asymptotically}, and algorithms like LinUCB~\citep{dani2008stochastic,chu2011contextual} play a central role in designing PROOF. Recent advances further improve the convergence rate under specific settings~\citep{bastani2020online,mintz2020nonstationary}.
Bandit data-driven optimization reduces to contextual bandit if we skip training an ML model and directly pick an action.
However, by doing so, we would effectively give up all the valuable historical data. Furthermore, although bandit algorithms have succeeded in millisecond-level decision-making~\citep{li2010contextual}, they are impractical in applications like food rescue where one time step represents a week, if not a month. The resulting long convergence time would hardly be acceptable to any stakeholders. We prove the same regret bound as previous work in our more realistic setting, with the regret decreasing much faster empirically.

Also related is offline policy learning~\citep{swaminathan2015batch,dudik2011doubly,athey2017efficient}. It does not need any online trials, and hence is much easier to convince the stakeholder to adopt. 
However, it assumes the historical data has various actions attempted, which fails to hold in most public sector applications.

In short, by proposing bandit data-driven optimization as a new learning paradigm, we fill a hole that no existing models were designed to address. 

\begin{table*}[t]
\label{tab:relatedwork}
\centering
\begin{tabular}{@{}lllll@{}}
\toprule
\textbf{\begin{tabular}[c]{@{}l@{}}Desired properties \end{tabular}}                & \textbf{\begin{tabular}[c]{@{}l@{}}Bandit  data-driven \\ optimization\end{tabular}}    & \begin{tabular}[c]{@{}l@{}}Data-driven \\ optimization\end{tabular}        & \begin{tabular}[c]{@{}l@{}}Contextual \\ bandit\end{tabular}                  & \begin{tabular}[c]{@{}l@{}}Offline \\ policy learning\end{tabular}     \\ \midrule
\textbf{\begin{tabular}[c]{@{}l@{}}No diverse past\\
data needed\end{tabular}}                       & \textbf{Yes}                                                                           & No                                                                         & Yes                                                                           & No                                                                            \\ \midrule
\textbf{\begin{tabular}[c]{@{}l@{}}Explicit learning\\ and  optimization\end{tabular}} & \textbf{Yes}                                                                           & Yes                                                                        & No                                                                            & No                                                                            \\ \midrule
\textbf{\begin{tabular}[c]{@{}l@{}}No assumption on \\ policy objective\end{tabular}}                       & \textbf{Yes}                                                                           & No                                                                         & \begin{tabular}[c]{@{}l@{}}Yes (but ignores \\ domain knowledge)\end{tabular} & \begin{tabular}[c]{@{}l@{}}Yes (but ignores \\ domain knowledge)\end{tabular} \\ \midrule
\textbf{\begin{tabular}[c]{@{}l@{}}Allows for \\iterative process\end{tabular}}                            & \textbf{Yes}                                                                           & No                                                                         & Yes                                                                           & Yes                                                                           \\ \midrule
\textbf{\begin{tabular}[c]{@{}l@{}}Finds optimal \\policy quickly\end{tabular}}                            & \textbf{\begin{tabular}[c]{@{}l@{}}Yes (compared\\ to  bandit)\end{tabular}} & \begin{tabular}[c]{@{}l@{}}Yes (if diverse \\ data available)\end{tabular} & No                                                                            & \begin{tabular}[c]{@{}l@{}}Yes (if diverse \\ data available)\end{tabular}    \\ \bottomrule
\end{tabular}

\caption{A comparison of different models regarding the desired properties in AI for non-profit and public sector applications.}
\end{table*}
\section{Bandit Data-Driven Optimization}
\label{sec:model}
We describe the formal setup of bandit data-driven optimization in Procedure~\ref{alg:procedure}. 
On Line~\ref{algline:dataset}, we receive an initial dataset $\mathcal D$ of size $n_0$, with features $x_i^{0}$ and label $c_i^{0}$ for data point $i$, and intervention in-place $w^{0}_i$ when the data point is collected.
Each feature vector $x_i^{0}$ is drawn i.i.d. from an unknown distribution $D_x$. Each label $c_i^{0}\in C$ is independently drawn from an unknown conditional distribution $D(w^{0}_i)_{c|x_i^{0}}$, which is parameterized by the intervention $w^{0}_i$, as different interventions could lead to different data distributions. 
In reality, $w^{0}_i$ is often identical across all $i$.
On Line~\ref{algline:ml}, we use all the data collected so far to train an ML model $f_t$, which is a mapping from features $X$ to labels $C$.
On Line~\ref{algline:opt}, we get a new set of feature samples $\mathbf{x}^t=\{x_i^t\}_{i=1}^{n}$. Then, we select an intervention $w_i^t \in W$ for each individual $i$. 
On Line~\ref{algline:adddata}, we commit to interventions $\mathbf{w}^t = \{w^t_i\}$ and receive the labels $\mathbf{c}^t=\{c_i^t\}$. Each label is independently drawn from the distribution $D(w^{t}_i)_{c|x_i^{t}}$. 
Then, on Line~\ref{algline:reward}, we incur a cost $u_t$.

We assume that the cost $u_t$ is determined by a partially known function $u(\mathbf{c}^t,\mathbf{w}^t)$. The function consists of three terms. The first term $\sum_i p(c_i^t, w^t_i)$ is the known loss. $p(c, w)$ is a fully known function capturing the loss for choosing intervention $w$ and getting label $c$. It represents our modeling effort and domain knowledge. The second term $\sum_i q(w^t_i)$ is the unknown loss. $q(w)$ is an unknown function representing all the unmodeled objectives and the unintended consequences of using the intervention $w$. The third term is random noise $\eta$.
This form of loss -- a known part $p(\cdot)$ and an unknown part $q(\cdot)$ -- is a realistic compromise of two extremes. We spend a lot of time communicating with food rescue practitioners to understand the problem. It would go against this honest effort to eliminate $p(\cdot)$ and model the process as a pure bandit problem. On the other hand, there will be unmodeled objectives, however hard we try. It would be too arrogant to eliminate $q(\cdot)$ and pretend that anything not going according to the plan is noise. 
The unknown $q(\cdot)$ is our acknowledgement that any intervention may have unintended consequences.
We leave to future work to consider other interactions between $p(\cdot)$ and $q(\cdot)$.

\begin{algorithm}[t]
	Receive initial dataset $\mathcal D = \{(x_i^{0}, c_i^{0}; w^{0}_i)_{i=1,\dots,n_0}\}$ from distribution $D$ on $(X, C)$.\\ \label{algline:dataset}
	\For{$t = 1, 2, \dots, T$}{
	Using all the available data $\mathcal D$, train ML prediction model $f_t: X \to C$.\\ \label{algline:ml}
    Given $n$ feature samples $\{x_i^t\} \sim D_x$, choose interventions $\{w^t_i\}$ for each individual $i$.\\\label{algline:opt}
	Receive $n$ labels $\{c_i^t\} \sim D(w^{t}_i)_{c|x_i^{t}}$. Add $\{(x_i^{t}, c_i^{t}; w^{t}_i)_{i=1,\dots,n}\}$ to the dataset $\mathcal D$.\\\label{algline:adddata}
	Get cost $u_t = u(\mathbf{c}^t, \mathbf{w}^t) = \sum_i p(c_i^t, w^t_i)$ $+ \sum_i q(w^t_i) + \eta$, where $\eta \sim N(0, \sigma^2)$.\\ \label{algline:reward}
	}
	\caption{\textsc{Bandit Data-driven Optimization}}
	\label{alg:procedure}
\end{algorithm}

Hence, the question is how to select the intervention $\mathbf{w}^t$.
As is typical in the bandit literature, we define the optimal 
policy to be that given feature $\mathbf{x}$, pick action $\pi(\mathbf{x})$ such that
\[
\pi(\mathbf{x}) = \arg\min_\mathbf{w} \mathbb E_{\mathbf{c}, \eta | \mathbf{x}} [u(\mathbf{c}, \mathbf{w})],
\]
where the expectation is taken over labels $\mathbf{c}$ and noise $\eta$ conditioned on the features $\mathbf{x}$.
The goal is to devise an algorithm to select interventions $\mathbf{w}^t$ to minimize the regret
\[
R_T =  \mathbb E_{x, c, \eta} \left[   \sum_{t=1}^T \left( u(\mathbf{c}^t, \mathbf{w}^t)   - u(\mathbf{c}^t, \pi(\mathbf{x}^t))  \right) \right].
\]

The label $c$ can be a scalar or a vector. For the rest of the paper, we assume $C\in \mathbb{R}^d$ and $W\in \mathbb{R}^d$. $W$ may be discrete or continuous but it is assumed to be bounded.

Bandit data-driven optimization could be applied to a variety of AI for non-profit and public sector projects across application domains. 
The canonical problem setting is the scarce resource allocation in the non-profit and public context where an intervention corresponds to a resource allocation plan and new interventions need to be chosen periodically based on the data collected under previous interventions.
For example, in game-theoretic anti-poaching, one trains an ML model using geospatial features to predict poacher activity, and then solves an optimization problem to find a patrol strategy~\citep{nguyen2016capture,fang2016deploying}. The patrol finds more poaching data points so we go back to update the ML model, starting another iteration of trial. In education programs, one trains an ML model to predict students' risk of dropping out, and then solves an optimization problem to allocate education resources to the students under budget and fairness constraints~\citep{lakkaraju2015machine}. After one round, one observes the students' performance and starts the next iteration of the program.
To illustrate how bandit data-driven optimization captures real-world AI for nonprofit workflows more concretely, we describe below one particular application, food rescue volunteer recommendation, in detail.

\subsection{Food Rescue Volunteer Recommendation as Bandit Data-Driven Optimization}
\label{sec:fr}
Wasted food account for 25\% of the US food consumption, while 12\% of the US population struggle with food insecurity~\citep{coleman2020household}. With the end of COVID-19 pandemic nowhere in sight, the problem is becoming even more serious~\citep{laborde2020covid}. From New York to Colorado, from San Francisco to Sydney, food rescue platforms are fighting against food waste and insecurity in over 100 cities around the world. Their operation has proved to be effective~\citep{wolfson2018savoring}. These platforms match food donations from restaurants and grocery stores to low-resource community organizations. 
Once this matching is done, the food rescue dispatcher would post the donor and recipient information on their mobile app. The volunteers will then receive push notifications about the rescue. They could then claim it on the app and then complete the rescue.

Relying on external volunteers brings great uncertainty to the food rescue operation. Occasionally, some rescue trips would stay unclaimed for a long time. Since unclaimed rescues would seriously discourage the donors and recipients from further participation, food rescue dispatchers want to prevent this as much as possible. 
The dispatcher may recommend each rescue to a subset of volunteers through push notifications, The selection of volunteers to notify is the intervention $w\in \{0,1\}^d$ (with the $j^{th}$ dimension representing whether to send notification to the $j^{th}$ volunteer). This decision is dependent on how likely a rescue will be claimed by each volunteer. Thus, we can use an ML-based recommender system which leverages the features of a rescue and the volunteers, e.g. donor/recipient location, weather, the volunteer's historical activities, etc. (feature $x$), to predict the probabilities that each volunteer will claim the rescue. Our previous work is focused exclusively on this static recommender system~\citep{shi2021recommender}. After we select $w$ for a rescue, we observe which volunteer actually claim the rescue, that is, the vector label $c$ (with the $j^{th}$ dimension representing whether the $j^{th}$ volunteer claims the rescue). This data point will be added to our dataset and used for training later. The base optimization objective $p(c, w)$ reflects the fact that we want to send notifications to the volunteers who will claim it, while not sending too many notifications. Obviously, whether or not the rescue gets claimed after these push notifications matter to the food rescue organization. 
Yet, there is more to the cost to the food rescue, e.g. how each volunteer reacts to push notifications (will they get annoyed and leave?). 
The $q(\cdot)$ cost could capture such factors. 
\section{Algorithms and Regret Analysis}
\label{sec:theory}
We propose a flexible algorithm for bandit data-driven optimization and establish a formal regret analysis.

The data points are drawn from $X \times C \subseteq \mathbb R^m \times \mathbb R^d$. We assume all $x \in X$ has $l^2$-norm bounded by constant $K_X$, and the label space $C$ has $l^1$-diameter $K_C$. The action space $W$ could be either discrete or continuous, but is bounded inside the unit $l^2$-ball in $\mathbb R^d$. We specify the data distribution by an arbitrary marginal distribution $D_x$ on $X$ and a conditional distribution such that $c = f(x) + \epsilon$ where $\epsilon \sim \mathcal N(0, \sigma^2 I)$, for some unknown function $f$. To begin with, we assume $f \in \mathcal F$ comes from the class of all linear functions with $f(x) = Fx$, and we use ordinary least squares regression as the learning algorithm. We will relax this assumption towards the end of Section~\ref{sec:unsmart}. 
The known cost $p(c, w) = c^\dagger w$ is the inner product of label $c$ and action $w$.\footnote{We use superscript $\dagger$ to denote matrix and vector transpose.} The unknown cost is $q(w) = \mu^\dagger w$, where $\mu$ is an unknown but fixed vector.
Furthermore, for exposition purpose we will start by assuming that the intervention $w$ does not affect the data distribution. In Section~\ref{sec:cf}, we will remove this assumption.

\subsection{With Exactly Known Objectives}
As a primer to our main results to be introduced in the following section, we first look into a special case where we know the optimization objective. That is, our cost only consists of $p(\cdot)$, with $q(\cdot) = 0$. This is not very realistic, but by studying it we will gain intuition for the general case.

At each iteration, this setting resembles the predict-then-optimize framework studied by~\citet{elmachtoub2017smart}. Given a sample feature $x$, we need to solve the linear program with a known feasible region $W \subseteq \mathbb R^d$:
\begin{align*}
    \min_w \quad & \mathbb E_{c \sim D_{c|x}} [p(c, w) | x] = \mathbb E_{c \sim D_{c|x}} [c | x]^\dagger w\\
    s.t. \quad & w \in W
\end{align*}
We hope to learn a predictor $\hat f: X \to C$ from the given dataset, so that we can solve the following problem instead.
\begin{align*}
   w^*(\hat c) :=  \arg\min_w \quad & \hat c^\dagger w \qquad \text{where} \quad \hat c = \hat f(x)\\
    s.t. \quad & w \in W
\end{align*}
In this paper we assume that the problem has a unique optimal solution.
Since the total cost is the same as the known optimization objective, intuitively we should simply commit to the action $w^*(\hat c)$. By doing so, the expected regret we incur on this data point is $\mathbb E_x [r(x)]$, where 
\[
r(x) = \mathbb E_{c|x}[c]^\dagger (w^*(\hat c) - w^*(\mathbb E_{c|x}[c])).
\]
Theorem~\ref{thm:q0regret} establishes that, indeed, this strategy leads to no-regret. This is not entirely trivial, because the optimization is based on the learned predictor yet the cost is based on the true distribution. The proof is instrumental to the subsequent results. All the proofs are in Appendix~\ref{app:proofs}.
\begin{theorem}
When the total cost is fully modeled, i.e. $q(\cdot) = 0$, simply following the predict-then-optimize optimal solution leads to regret $O(\sqrt{ndmT})$.
\label{thm:q0regret}
\end{theorem}

\subsection{PROOF: Predict-then-Optimize with Optimism in Face of Uncertainty}
\label{sec:unsmart}
When there is no bandit uncertainty, as we showed just now one can simply follow the predict-then-optimize framework and no-regret is guaranteed. However, the unknown bandit cost is crucial to real-world food rescue and similar applications. We now describe the first algorithm for bandit data-driven optimization, PRedict-then-Optimize with Optimism in Face of uncertainty (PROOF), shown in Algorithm~\ref{alg:p2pofu}.

\SetAlgorithmName{Algorithm}{algorithm}{List of Algorithms}

\begin{algorithm}[t]
\KwInitialize{Find a barycentric spanner $b_1,\dots, b_d$ for $W$\\
    Set $A^1_i = \sum_{j=1}^d b_j b_j^\dagger$ and $\hat \mu^1_i = 0$ for $i = 1,2,\dots, n$.\\
}
Receive initial dataset $\mathcal D = \{(x_i^{0}, c_i^{0}; w^{0}_i)_{i=1,\dots,n_0}\}$ from distribution $D$ on $(X, C)$.\\
	\For{$t = 1, 2, \dots, T$}{
		Using all data in $\mathcal D$, train ML model $f_t: X \to C$. \\ 
		Given $n$ feature samples $\{x_i^t\} \sim D_x$, get predictions $\hat c_i^t = f_t(x^t_i)$.\\
		Set $\beta^t = \max\left( 128 d\log t \log\frac{n t^2}{\gamma}, \left( \frac{8}{3} \log \frac{n t^2}{\gamma} \right)^2 \right)$\\
		\For{$i=1,2,\dots, n$}{
	        Confidence ball $B_i^t = \{\nu: ||\nu - \hat \mu^t_i||_{2, A^t_i} \leq \sqrt{\beta^t} \}$.\\
		Choose intervention $w^t_i = \arg\min_{w\in W} \min_{\nu \in B^t_i} (\hat c^t_i + \nu)^\dagger w$. \\ \label{algline:merge}
		Receive label $c_i^t \sim D_{c|x_i^t}$. Add $(x_i^t, c_i^t; w_i^t)$ to $\mathcal D$. \\
		 Get cost $u^t_i = u(x^t_i, c_i^t, w^t_i) = (c^t_i)^\dagger w_i^t $ $+ \mu^\dagger w^t_i + \eta_i$, where $\eta_i \sim N(0, \sigma^2)$. Let $u^t_{oi}$ be the 1st term and let $u^t_{bi}$ be the sum of the 2nd and 3rd term.\\ 
		Update $A^{t+1}_i = A^t_i + w^t_i (w^t_i)^\dagger$ \\ \label{algline:updateA}
		Update $\hat \mu^{t+1}_i = (A^{t+1}_i)^{-1} \sum_{\tau=1}^t u^t_{bi} w^t_i$ \label{algline:updatemu}
		}
	}
	\caption{\textsc{PROOF: Predict-then-optimize with optimism in face of uncertainty}}\label{alg:p2pofu}
\end{algorithm}

PROOF is an integration of the celebrated Optimism in Face of Uncertainty (OFU) framework and the predict-then-optimize framework.
It is clear that the unknown cost component $q(\cdot) + \eta$ forms a linear bandit.
For this bandit component, we run an OFU algorithm for each individual $i$ with the same unknown loss vector $\mu$. The OFU component for each individual $i$ maintains a confidence ball $B^t_i$ which is independent of the predict-optimize framework. The predict-then-optimize framework produces an estimated optimization objective $\hat c^t$ independent of OFU. The two components are integrated together on Line~\ref{algline:merge} of Algorithm~\ref{alg:p2pofu}, where we compute the intervention for the current round taking into consideration the essence of both frameworks.

Below, we justify why this algorithm achieves no-regret.
First, we state a theorem by~\citet{dani2008stochastic}, which states that the confidence ball captures the true loss vector $\mu$ with high probability. The result was proved for the original OFU algorithm. However, since the result itself does not depend on the way we choose $w^t$, it still holds in bandit data-driven optimization.
We adapt it by adding a union bound so that the result holds for all the $n$ bandits simultaneously.
\begin{lemma}[Adapted from Theorem 5 by~\citet{dani2008stochastic}]
\label{lem:confidenceball} 
Let $\gamma > 0$, then
$\mathbb P( \forall t, \forall i, \mu \in B^t_i) \geq 1-\gamma$.
\end{lemma}
The following key lemma decomposes the regret into two components: one involving the online bandit loss, the other concerning the offline supervised learning loss.

\begin{lemma}
With probability $1-\delta$, Algorithm~\ref{alg:p2pofu} has regret
\[
O \left( n\sqrt{8m T \beta^T \log T} + \sum_{t=1}^T \sum_{i=1}^n  \mathbb E \left[ \left\| \mathbb E_{c_i^t | x_i^t}[c_i^t] - \hat c^t_i \right\|_2 \right] \right).
\]
\label{lem:regret}
\end{lemma}

Clearly, to characterize the regret, we need to bound $\mathbb E \left[ \left\|  \mathbb E_{c_i^t | x_i^t}[c_i^t] - \hat c^t_i \right\|_2 \right]$. In the case of linear regression, we have the following theorem.

\begin{theorem}
Assuming we use ordinary least squares regression as the ML algorithm, Algorithm~\ref{alg:p2pofu} has regret $\tilde O\left( n\sqrt{dmT} \right)$ with probability $1-\delta$.
\label{thm:regret_linear}
\end{theorem}

Theorem~\ref{thm:regret_linear} assumes a linear regression problem with a specific learning algorithm -- ordinary least squares linear regression. If our intent is for Algorithm~\ref{alg:p2pofu} to be modular where one can use any learning algorithm, we could resort to sample complexity bounds. In Appendix~\ref{app:rad}, we include a derivation of the regret bound from the sample complexity perspective. This approach allows us to extend the result in Theorem~\ref{thm:regret_linear} to a more general setting.

\subsection{When Interventions Affect Label Distribution}
\label{sec:cf}
So far in this section, we have had the assumption that the action $w$ does not affect the distribution $D$ from which as sample $(X, C)$. In many real-world scenarios this is not the case. For example, if the wildlife patrollers change their patrol routes, the poachers' poaching location would change accordingly and hence its distribution would be very different. Thus, it is valuable to study this more general setting where the intervention could affect the label distribution.

First, let us make the assumption that there are finitely many possible actions. We will consider the continuous action space later.
Since there are finitely many actions, an intuitive idea is to train an ML predictor for each action separately. Because we do not impose any assumption on our initial dataset, which might only have a single action embedded, we clearly need to use exploration in the bandit algorithm and use the data points gathered along the way to train the predictor. It might seem very natural to fit this directly into the framework of PROOF as shown in Algorithm~\ref{alg:p2pofu}: simply maintain several predictors instead of one, and still choose the best action on Line~\ref{algline:merge}. However, to train the predictor corresponding to each action, we need at least a certain number of data points to bound the prediction error. Yet, PROOF, and UCB-type algorithms in general, do not give a lower bound on how many times each action is tried. For example, Algorithm~\ref{alg:p2pofu} might never try some action at all, and we would not be able to train a predictor for that action. To resolve this philosophical contradiction, we add a uniform exploration phase of length $\tilde T$ at the beginning, where at each round $1,2,\dots, \tilde T$, each action is taken on some examples. 
Other than this, we inherit all the setup for the analysis in Section~\ref{sec:unsmart}. We describe the detailed procedure as Algorithm~\ref{alg:p2pofucf} in Appendix~\ref{app:algo}.

We establish the following lemma which decomposes the regret into 3 parts: regret during uniform exploration, regret in UCB bandit, and regret through supervised learning.

\begin{lemma}
With probability $1-\delta$, Algorithm~\ref{alg:p2pofucf} has regret
\[
\begin{split}
O&\biggl( n\tilde T + n\sqrt{8mT \beta_T \log T} \\
&+ \sum_{t=\tilde T + 1}^T \sum_{i=1}^n \mathbb E \left[ \left\| \mathbb E_{c_i^t | x_i^t, w_i^t} [c^t_i(w_i^t)] - \hat c^t_i(w^t_i) \right\|_2 \right]  \biggr).
\end{split}
\]
\label{lem:regret_cf}
\end{lemma}

By combining Lemma~\ref{lem:regret_cf} with previous results, we arrive at the regret of PROOF in this more general setting.
\begin{theorem}
With finitely many actions and OLS as the ML algorithm, Algorithm~\ref{alg:p2pofucf} has regret $\tilde O\left( n (d |W|)^{1/3} m^{1/2} T^{2/3}  \right)$.
\label{thm:regret_linear_cf}
\end{theorem}

\begin{figure*}
    \subfloat[Small scale base case]{\includegraphics[width=0.33\columnwidth]{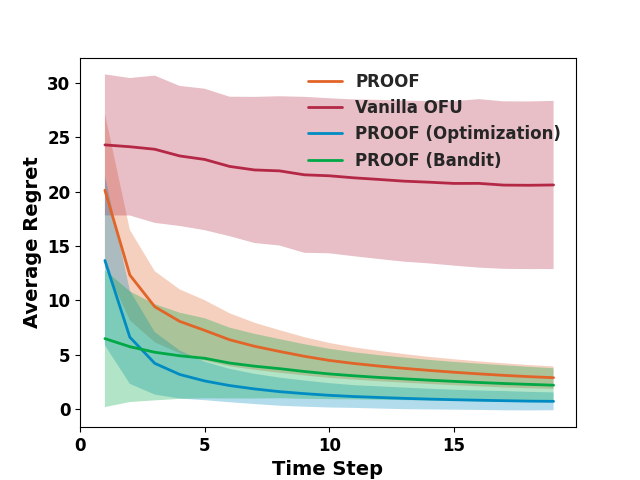}\quad
    \label{fig:n20m20d5ka10eta0.0001eps0.1}}
    \subfloat[Data per step increased from 20 to 40]{\includegraphics[width=0.33\columnwidth]{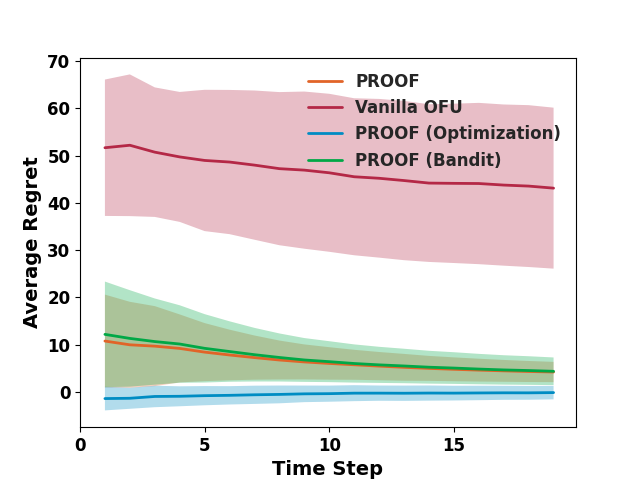}\quad
    \label{fig:n40m20d5ka10eta0.0001eps0.1}} 
    \subfloat[Linear mapping norm multiplied by 10.]{\includegraphics[width=0.33\columnwidth]{figures/n20m20d5ka100eta00001eps01iter500algOLSFINAL10.png}
    \label{fig:n20m20d5ka100eta0.0001eps0.1}} 
  \\ 
    \subfloat[Large scale base case.]{\includegraphics[width=0.33\columnwidth]{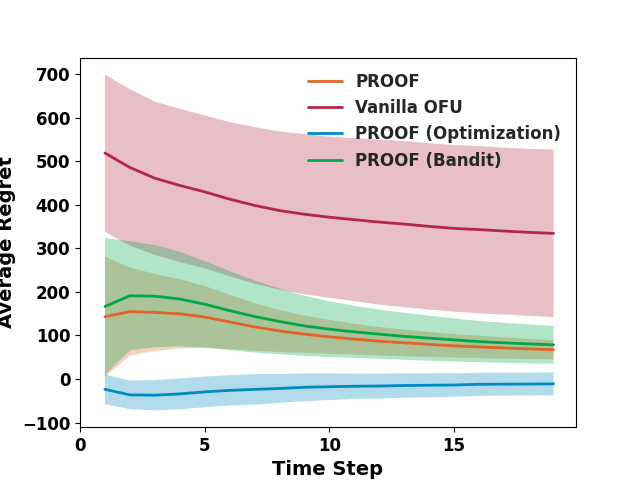} \quad
    \label{fig:n500m50d5ka10eta0.0001eps0.1}}
    \subfloat[Linear mapping norm divided by 10.]{\includegraphics[width=0.33\columnwidth]{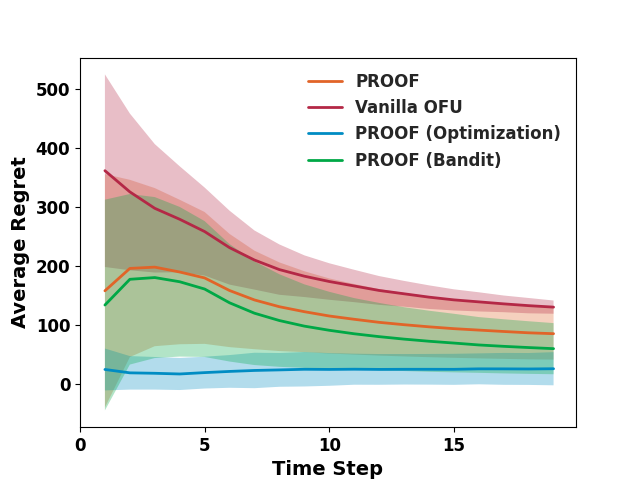}\quad
    \label{fig:n500m50d5ka1eta0.0001eps0.1}}
    \subfloat[Data noise multiplied by 5.]{\includegraphics[width=0.33\columnwidth]{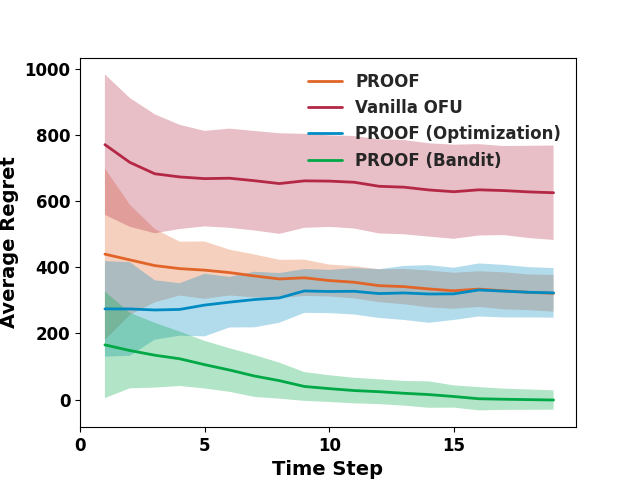}
    \label{fig:n500m50d5ka10eta0.0001eps0.5}}
    \caption{Numerical simulation results of PROOF compared against vanilla linear bandit. All results are averaged over 10 runs with shaded areas representing the standard deviation.}
    \label{fig:all}
\end{figure*}

We now move on to the scenario where the action space $W$ is continuous. In this case, we assume the true label of feature $x$ under action $w$ is $c = Fx + Gw + \epsilon$ where $\epsilon \sim \mathcal N(0, \sigma^2 I)$. A small modification of Algorithm~\ref{alg:p2pofucf} will work in this scenario: instead of rotating over each action in the uniform exploration phase, we simply pick action $w$ uniformly at random for each individual. Then, the regret of the algorithm is as follows.

\begin{theorem}
\label{thm:regret_linear_cf_cont}
Suppose the action space is continuous and the label can be modeled as a linear function of the feature and action. Assuming OLS as the ML algorithm, Algorithm~\ref{alg:p2pofucf} has regret $\tilde O\left( m^{1/3} d^{2/3} n T^{2/3} \right)$.
\end{theorem}

\subsection{PROOF Is a Modular Algorithm}
In practice, PROOF can be applied beyond the setting under which we proved the previous results. Rather than a fixed algorithm, it is designed to be modular so that we can plug in different learning algorithms and optimization problems. 
First, instead of linear regression, PROOF can accommodate any predictive model such as tree-based models and neural networks. Second, The nominal optimization problem need not be a linear optimization problem. The optimization problem may be continuous or discrete, convex or non-convex, as we do not concern ourselves with computational complexity in this paper.
In Section~\ref{sec:frexperiment}, we demonstrate that even when we insert complex algorithms into the PROOF framework, thereby going beyond the setting where we established formal regret guarantees, PROOF still works well.

\section{Experimental Results}
\subsection{Numerical Simulations}
\label{sec:expunsmart}

\begin{figure*}[t]
\centering
    \subfloat[Base case]{\includegraphics[width=0.33\columnwidth]{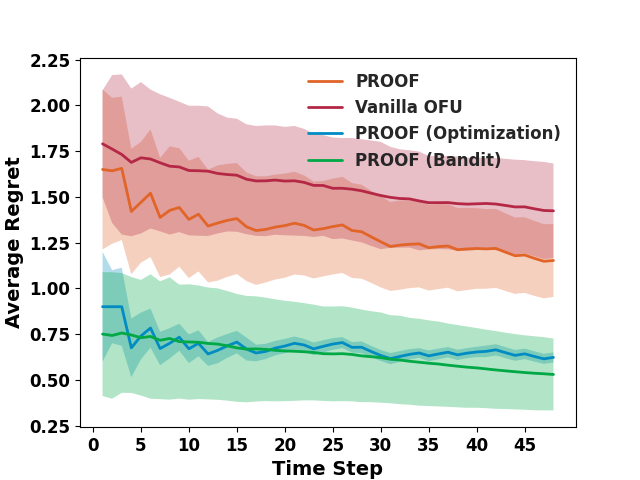}
    \label{fig:m6d100init300valid150top10rs1pos1}}
    \subfloat[Known cost $p(\cdot)$ multiplied by 4]{\includegraphics[width=0.33\columnwidth]{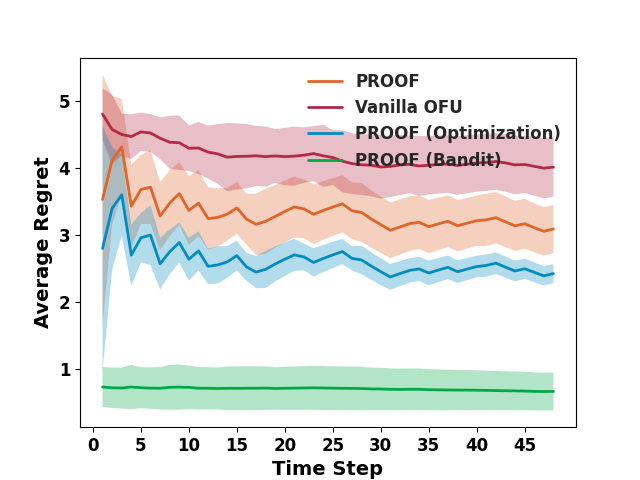}
    \label{fig:m6d100init300valid150top10rs0.25pos1}}
    \subfloat[Initial dataset size decreased to 20]{\includegraphics[width=0.33\columnwidth]{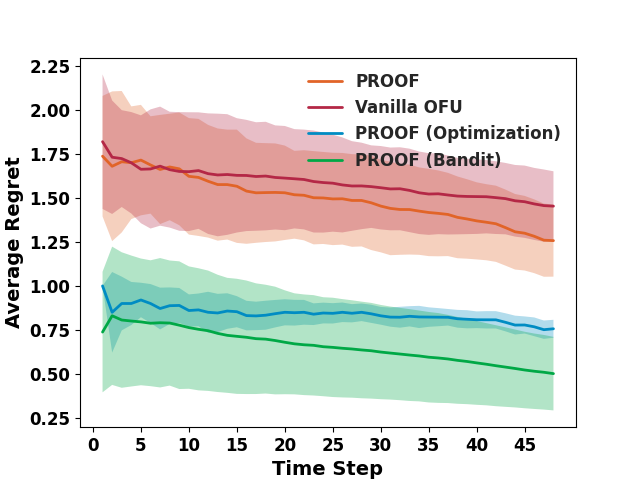}
    \label{fig:m6d100init20valid150top10rs1pos1}}
    \caption{The experiment results on the real-world food rescue data of PROOF compared against vanilla linear bandit. All results are averaged over 10 runs with shaded areas representing the standard deviation.}
    \label{fig:fr}
\end{figure*}

As the first step of validation, we implement PROOF in the setting described in Section~\ref{sec:unsmart} on a simulated dataset. 
We start with a small-scale experiment.
Recall that we train an ML predictor $\hat f: X \to C$ where $X \subseteq \mathbb R^m$ and $C \subseteq \mathbb R^d$. 
We take feature dimension $m=20$ and label dimension $d=5$. At every round we get $n=20$ data points. As is typical, we assume the bandit reward is bounded in $[-1,1]$ and the feasible region $W$ is the unit $l_2$-ball. For the true linear map $F$ where $c = Fx + \epsilon$, we upper bound its $l_1$ matrix norm at 10. We sample the noise $\epsilon \sim \mathcal N(0, \sigma^2 I_d)$ from a normal distribution where $\sigma^2 = 0.1$. We take the bandit noise $\eta \sim N(0, 10^{-4})$. We use OLS at each time step. We solve the non-convex program on Line~\ref{algline:merge} in Algorithm~\ref{alg:p2pofu} with IPOPT. We find the best action given the true reward parameters using Gurobi. 
We set $\beta^t = 1$ so that the algorithm can quickly concentrate on the region of interest.\footnote{The code for all the experiments in this section is available at \url{https://github.com/AIandSocialGoodLab/bandit-data-driven-optimization}}

The expected cost for a fixed action $w$ is $\mathbb E_{c, \eta}[(c+\mu)^\dagger w + \eta] = \mathbb E[x^\dagger F^\dagger w] + \mu^\dagger w  = \mu^\dagger w$,
because when we generated $x$, the distribution has zero mean. This problem in theory might be solved as a linear bandit by feeding the total cost to OFU. Since the regret bound of OFU is the same as PROOF in the order of $T$, this brings back the point that we have been emphasizing since the beginning: if linear bandit is a more general model whose algorithms already solve our problem, why would we care about bandit data-driven optimization? 

In Sections~\ref{sec:intro} through~\ref{sec:model}, we answered this question with the characteristics of the food rescue. Here, we answer this question using experiments.
We show the average regret of PROOF as the orange curve in Figure~\ref{fig:all}, and that of OFU in red. 
We can decompose the average regret of PROOF into the regret of the optimization component and the regret of the bandit component. The former is simply the algorithm's optimization (known) cost minus the best intervention's optimization cost. The latter is defined similarly. Neither needs to be positive.
Figure~\ref{fig:n20m20d5ka10eta0.0001eps0.1} shows that PROOF quickly reduces the regret in both components, while the performance of vanilla OFU is much more underwhelming. 
This difference is because an offline predictive model captures the large variance in the implicit context $x$ and $c$ much better.
In fact, PROOF consistently has much smaller variance than OFU. 

We now tweak the parameters a bit. When we increase the number of data points per iteration from $n=20$ to 40, Figure~\ref{fig:n40m20d5ka10eta0.0001eps0.1} shows that the regret of the optimization component becomes very small to start with, because we have more data to learn from. When we increase $||F||$ from 10 to 100, Figure~\ref{fig:n20m20d5ka100eta0.0001eps0.1} shows that the optimization regret dominates the total regret, as the optimization cost is now much larger than the bandit cost. Here, vanilla OFU suffers even more, because now its cost signal has even larger magnitude and variance. 

We then scale up the experiments and show that PROOF still outperforms OFU even when the problem parameters are not as friendly. Suppose we get $n=500$ data points every time and each data point has $m=50$ features. Keeping all other parameters unchanged, Fig.~\ref{fig:n500m50d5ka10eta0.0001eps0.1} shows that PROOF still outperforms OFU by a lot. 
In Fig.~\ref{fig:n500m50d5ka1eta0.0001eps0.1}, we change $||F||$ from 10 to 1, making the optimization cost less important. This reduces the variance of OFU and it is doing better than previously. However, our PROOF still outperforms OFU. In Fig.~\ref{fig:n500m50d5ka10eta0.0001eps0.5}, we increase the label noise from $\epsilon \sim \mathcal N(0, 0.1I_d)$ to $\mathcal N(0, 0.5I_d)$. This poses more challenge to PROOF. 
But still, PROOF manages to keep its regret below OFU.

\subsection{Food Rescue Volunteer Recommendation}
\label{sec:frexperiment}

Bandit data-driven optimization is motivated by the practical challenges in the deployment of AI for non-profit and public sector projects. After abstracting these challenges to a theoretical model, we now return to the food rescue volunteer recommendation problem. We have introduced the details of food rescue operations in Section~\ref{sec:fr}.

There are 100 volunteers. At each time step, we get a new rescue and decide a subset of 10 volunteers to whom we send push notifications. We represent this action with a binary vector $w \in \{0,1\}^{100}$ such that $w_i = 1$ if volunteer $i$ is notified and $0$ otherwise. Thus, the feasible action space $W$ is $\{0,1\}^{100}$ with the constraint of $\sum_{i=1}^{100} w_i \leq 10$. 
The action $w$ we take at each time step is backed by a content-based ML recommender system. The ML model receives a feature vector $x$ which describes a particular rescue-volunteer pair, and outputs a label prediction $\hat c$ as the likelihood of this volunteer claiming this rescue.\footnote{Here the label is 1-dimensional while our action space is 100-dimensional. This is easy to resolve. Each rescue-volunteer pair has $m'$ features. While in practice we have $\hat f': \mathbb R^{m'} \to \mathbb R$ and pass 100 feature vectors to it serially, one could think of a product model $\hat f = \prod_{i=1}^{100} \hat f'$ which takes the concatenation of 100 feature vectors and outputs a 100-dimensional vector.} 
We adapt this ML component from the one studied in~\citep{shi2021recommender}. Its feature selection, model architecture, and training techniques, are not trivial. Yet, since they are not the focus of this paper, we include all these details in Appendix~\ref{app:fr}.
The actual label $c$ is a one-hot vector in $\{0,1\}^{100}$ indicating which volunteer actually claimed the rescue.
The known cost $p(c, w) = c^\dagger w$ is 1 if we notify a volunteer who eventually claimed a rescue and 0 otherwise. To minimize it, we could negate the label $c$ (and its prediction $\hat c$). The bandit cost $q(\cdot)$ is the same as before. We solve the optimization at each time step of PROOF with Gurobi after applying a standard linearization trick~\citep{liberti2009reformulations}. We also gradually decrease the confidence radius $\beta$.

Unlike the case in Section~\ref{sec:expunsmart}, OFU algorithm does not work here in principle. This is because, working with real-world data, we do not know the data distribution and it is almost certainly not zero-mean. In fact, this experiment has also gone beyond the setting for which we proved formal regret bound for PROOF, yet we would like to see how these two algorithms perform in such a real-world use case.

We assume an initial dataset of 300 rescues and run the algorithms for 50 time steps, each time step corresponding to one new rescue. As shown in Figure~\ref{fig:m6d100init300valid150top10rs1pos1}, PROOF outperforms vanilla OFU by roughly 15\%. The performance gain by PROOF can be contributed to its effective use of the available data, as the progress on bandit made by PROOF and vanilla OFU are quite similar. In Figure~\ref{fig:m6d100init300valid150top10rs0.25pos1}, we scale up the known part of the cost by a factor of 4. Because the optimization is more emphasized, it is unsurprising to see that most of PROOF's progress depends on the recommender system itself. In this case, it has a larger performance margin over vanilla OFU. Finally, in Figure~\ref{fig:m6d100init20valid150top10rs1pos1} we decrease the size of the initial dataset from 300 rescues to 20 rescues. We observe that PROOF still has an edge over vanilla OFU. The margin is minimal at the initial time steps, because we have much less initial information here. Yet still, as time goes by PROOF picks up more information in the feature/label dataset to expand its margin. In actual food rescue projects, the amount of initial data is typically more than this, more resembling Figure~\ref{fig:m6d100init300valid150top10rs1pos1}, but Figure~\ref{fig:m6d100init20valid150top10rs1pos1} assures us that PROOF still works in this more extreme case.

\section{Conclusion}
Non-profit and public sectors have huge potential to benefit from the advancing machine learning research. However, plenty of experience shows that the machine learning model itself is almost always not enough to address the real-world societal challenges. Motivated by four practical pain points in such applications, we proposed bandit data-driven optimization, designed the PROOF algorithm, and showed that it has no-regret. Finally, we show its better performance over bandit algorithm in simulations and the food rescue context.
We view bandit data-driven optimization as our first attempt to bridge the last-mile gap between static ML models and their actual deployment in the real-world non-profit and public context. 
\section*{Acknowledgments}
We have learned so much from our collaborators at 412 Food Rescue and other nonprofit organizations and we thank them for what they are doing for our community.
We thank Thomas G. Dietterich, Fatma Kılınç-Karzan and Kit T. Rodolfa for the inspiring discussions at the early stage of this work.
This work was supported in part by NSF grants IIS-1850477 and IIS-2046640 (CAREER), a Siebel Scholarship and a Carnegie Mellon Presidential Fellowship. The views and conclusions contained in this document are those of the authors and should not be interpreted as representing the official policies, either expressed or implied, of the funding agencies.
\bibliography{ref}

\begin{thebibliography}{33}
\providecommand{\natexlab}[1]{#1}
\providecommand{\url}[1]{\texttt{#1}}
\expandafter\ifx\csname urlstyle\endcsname\relax
  \providecommand{\doi}[1]{doi: #1}\else
  \providecommand{\doi}{doi: \begingroup \urlstyle{rm}\Url}\fi

\bibitem[Abernethy et~al.(2018)Abernethy, Chojnacki, Farahi, Schwartz, and
  Webb]{abernethy2018activeremediation}
Jacob Abernethy, Alex Chojnacki, Arya Farahi, Eric Schwartz, and Jared Webb.
\newblock Activeremediation: The search for lead pipes in flint, michigan.
\newblock In \emph{KDD}, pages 5--14, 2018.

\bibitem[Apostol(1966)]{apostolintroduction}
Tom~M Apostol.
\newblock Introduction to analytic number theory.
\newblock 1966.

\bibitem[Athey and Wager(2017)]{athey2017efficient}
Susan Athey and Stefan Wager.
\newblock Efficient policy learning.
\newblock \emph{arXiv:1702.02896}, 2017.

\bibitem[Balkanski et~al.(2016)Balkanski, Rubinstein, and
  Singer]{balkanski2016power}
Eric Balkanski, Aviad Rubinstein, and Yaron Singer.
\newblock The power of optimization from samples.
\newblock In \emph{NIPS}, pages 4017--4025, 2016.

\bibitem[Ban and Rudin(2019)]{ban2019big}
Gah-Yi Ban and Cynthia Rudin.
\newblock The big data newsvendor: Practical insights from machine learning.
\newblock \emph{Operations Research}, 67\penalty0 (1):\penalty0 90--108, 2019.

\bibitem[Bastani and Bayati(2020)]{bastani2020online}
Hamsa Bastani and Mohsen Bayati.
\newblock Online decision making with high-dimensional covariates.
\newblock \emph{Operations Research}, 2020.

\bibitem[Bertsimas and Kallus(2020)]{bertsimas2020predictive}
Dimitris Bertsimas and Nathan Kallus.
\newblock From predictive to prescriptive analytics.
\newblock \emph{Management Science}, 66\penalty0 (3):\penalty0 1025--1044,
  2020.

\bibitem[Chen et~al.(2017)Chen, Gupta, Li, Qiao, and Wang]{chen2017nearly}
Lijie Chen, Anupam Gupta, Jian Li, Mingda Qiao, and Ruosong Wang.
\newblock Nearly optimal sampling algorithms for combinatorial pure
  exploration.
\newblock In \emph{Conference on Learning Theory}, pages 482--534. PMLR, 2017.

\bibitem[Chu et~al.(2011)Chu, Li, Reyzin, and Schapire]{chu2011contextual}
Wei Chu, Lihong Li, Lev Reyzin, and Robert Schapire.
\newblock Contextual bandits with linear payoff functions.
\newblock In \emph{Proceedings of the Fourteenth International Conference on
  Artificial Intelligence and Statistics}, pages 208--214, 2011.

\bibitem[Coleman-Jensen et~al.(2020)Coleman-Jensen, Rabbitt, Gregory, and
  Singh]{coleman2020household}
Alisha Coleman-Jensen, Matthew~P Rabbitt, Christian~A Gregory, and Anita Singh.
\newblock Household food security in the united states in 2019.
\newblock \emph{USDA-ERS Economic Research Report}, 2020.

\bibitem[Dani et~al.(2008)Dani, Hayes, and Kakade]{dani2008stochastic}
Varsha Dani, Thomas~P Hayes, and Sham~M Kakade.
\newblock Stochastic linear optimization under bandit feedback.
\newblock In \emph{Conference on Learning Theory}, 2008.

\bibitem[Dong et~al.(2018)Dong, Chen, and Zeng]{dong2018generalized}
Chaosheng Dong, Yiran Chen, and Bo~Zeng.
\newblock Generalized inverse optimization through online learning.
\newblock In \emph{NeurIPS}, 2018.

\bibitem[Dud{\'\i}k et~al.(2011)Dud{\'\i}k, Langford, and Li]{dudik2011doubly}
Miroslav Dud{\'\i}k, John Langford, and Lihong Li.
\newblock Doubly robust policy evaluation and learning.
\newblock \emph{arXiv preprint arXiv:1103.4601}, 2011.

\bibitem[Elmachtoub and Grigas(2017)]{elmachtoub2017smart}
Adam~N Elmachtoub and Paul Grigas.
\newblock Smart" predict, then optimize".
\newblock \emph{arXiv preprint arXiv:1710.08005}, 2017.

\bibitem[Esfahani et~al.(2018)Esfahani, Shafieezadeh-Abadeh, Hanasusanto, and
  Kuhn]{esfahani2018data}
Peyman~Mohajerin Esfahani, Soroosh Shafieezadeh-Abadeh, Grani~A Hanasusanto,
  and Daniel Kuhn.
\newblock Data-driven inverse optimization with imperfect information.
\newblock \emph{Mathematical Programming}, 167\penalty0 (1):\penalty0 191--234,
  2018.

\bibitem[Fang et~al.(2016)Fang, Nguyen, Pickles, Lam, Clements, An, Singh,
  Tambe, and Lemieux]{fang2016deploying}
Fei Fang, Thanh~H Nguyen, Rob Pickles, Wai~Y Lam, Gopalasamy~R Clements, Bo~An,
  Amandeep Singh, Milind Tambe, and Andrew Lemieux.
\newblock Deploying paws: Field optimization of the protection assistant for
  wildlife security.
\newblock In \emph{Twenty-eighth IAAI conference}, 2016.

\bibitem[Gholami et~al.(2019)Gholami, Yadav, Tran-Thanh, Dilkina, and
  Tambe]{gholami2019don}
Shahrzad Gholami, Amulya Yadav, Long Tran-Thanh, Bistra Dilkina, and Milind
  Tambe.
\newblock Don't put all your strategies in one basket: Playing green security
  games with imperfect prior knowledge.
\newblock In \emph{AAMAS}, pages 395--403, 2019.

\bibitem[Golowich et~al.(2018)Golowich, Rakhlin, and Shamir]{golowich2018size}
Noah Golowich, Alexander Rakhlin, and Ohad Shamir.
\newblock Size-independent sample complexity of neural networks.
\newblock In \emph{Conference On Learning Theory}, pages 297--299. PMLR, 2018.

\bibitem[Kao et~al.(2009)Kao, Roy, and Yan]{kao2009directed}
Yi-hao Kao, Benjamin~V Roy, and Xiang Yan.
\newblock Directed regression.
\newblock In \emph{Advances in Neural Information Processing Systems}, pages
  889--897, 2009.

\bibitem[Laborde et~al.(2020)Laborde, Martin, Swinnen, and
  Vos]{laborde2020covid}
David Laborde, Will Martin, Johan Swinnen, and Rob Vos.
\newblock Covid-19 risks to global food security.
\newblock \emph{Science}, 369\penalty0 (6503):\penalty0 500--502, 2020.

\bibitem[Lai and Robbins(1985)]{lai1985asymptotically}
Tze~Leung Lai and Herbert Robbins.
\newblock Asymptotically efficient adaptive allocation rules.
\newblock \emph{Advances in applied mathematics}, 6\penalty0 (1):\penalty0
  4--22, 1985.

\bibitem[Lakkaraju et~al.(2015)Lakkaraju, Aguiar, Shan, Miller, Bhanpuri,
  Ghani, and Addison]{lakkaraju2015machine}
Himabindu Lakkaraju, Everaldo Aguiar, Carl Shan, David Miller, Nasir Bhanpuri,
  Rayid Ghani, and Kecia~L Addison.
\newblock A machine learning framework to identify students at risk of adverse
  academic outcomes.
\newblock In \emph{KDD}, pages 1909--1918, 2015.

\bibitem[Li et~al.(2010)Li, Chu, Langford, and Schapire]{li2010contextual}
Lihong Li, Wei Chu, John Langford, and Robert~E Schapire.
\newblock A contextual-bandit approach to personalized news article
  recommendation.
\newblock In \emph{WWW}, pages 661--670, 2010.

\bibitem[Liberti et~al.(2009)Liberti, Cafieri, and
  Tarissan]{liberti2009reformulations}
Leo Liberti, Sonia Cafieri, and Fabien Tarissan.
\newblock Reformulations in mathematical programming: A computational approach.
\newblock In \emph{Foundations of Computational Intelligence Volume 3}, pages
  153--234. Springer, 2009.

\bibitem[Mintz et~al.(2020)Mintz, Aswani, Kaminsky, Flowers, and
  Fukuoka]{mintz2020nonstationary}
Yonatan Mintz, Anil Aswani, Philip Kaminsky, Elena Flowers, and Yoshimi
  Fukuoka.
\newblock Nonstationary bandits with habituation and recovery dynamics.
\newblock \emph{Operations Research}, 68\penalty0 (5):\penalty0 1493--1516,
  2020.

\bibitem[Nguyen et~al.(2016)Nguyen, Sinha, Gholami, Plumptre, Joppa, Tambe,
  Driciru, Wanyama, Rwetsiba, Critchlow, et~al.]{nguyen2016capture}
Thanh~H Nguyen, Arunesh Sinha, Shahrzad Gholami, Andrew Plumptre, Lucas Joppa,
  Milind Tambe, Margaret Driciru, Fred Wanyama, Aggrey Rwetsiba, Rob Critchlow,
  et~al.
\newblock Capture: A new predictive anti-poaching tool for wildlife protection.
\newblock In \emph{AAMAS}, pages 767--775, 2016.

\bibitem[Perrault et~al.(2020)Perrault, Fang, Sinha, and Tambe]{perrault2019ai}
Andrew Perrault, Fei Fang, Arunesh Sinha, and Milind Tambe.
\newblock Ai for social impact: Learning and planning in the data-to-deployment
  pipeline.
\newblock \emph{AI Magazine}, 41\penalty0 (4):\penalty0 3--16, 2020.

\bibitem[Rosset and Tibshirani(2020)]{rosset2020fixed}
Saharon Rosset and Ryan~J Tibshirani.
\newblock From fixed-x to random-x regression: Bias-variance decompositions,
  covariance penalties, and prediction error estimation.
\newblock \emph{Journal of the American Statistical Association}, 115\penalty0
  (529):\penalty0 138--151, 2020.

\bibitem[{Shi} et~al.(2020){Shi}, {Yuan}, {Lo}, {Lizarondo}, and
  {Fang}]{shi2020improving}
Zheyuan~Ryan {Shi}, Yiwen {Yuan}, Kimberly {Lo}, Leah {Lizarondo}, and Fei
  {Fang}.
\newblock Improving efficiency of volunteer-based food rescue operations.
\newblock \emph{Proceedings of the AAAI Conference on Artificial Intelligence},
  34\penalty0 (8):\penalty0 13369--13375, 2020.

\bibitem[Shi et~al.(2021)Shi, Lizarondo, and Fang]{shi2021recommender}
Zheyuan~Ryan Shi, Leah Lizarondo, and Fei Fang.
\newblock A recommender system for crowdsourcing food rescue platforms.
\newblock In \emph{Proceedings of the Web Conference 2021}, pages 857--865,
  2021.

\bibitem[Swaminathan and Joachims(2015)]{swaminathan2015batch}
Adith Swaminathan and Thorsten Joachims.
\newblock Batch learning from logged bandit feedback through counterfactual
  risk minimization.
\newblock \emph{JMLR}, 16\penalty0 (1):\penalty0 1731--1755, 2015.

\bibitem[Wolfson and Greeno(2018)]{wolfson2018savoring}
Megan~D Wolfson and Catherine Greeno.
\newblock Savoring surplus: effects of food rescue on recipients.
\newblock \emph{Journal of Hunger \& Environmental Nutrition}, 2018.

\bibitem[Zheng et~al.(2018)Zheng, Waggoner, Liu, and Chen]{zheng2018active}
Shuran Zheng, Bo~Waggoner, Yang Liu, and Yiling Chen.
\newblock Active information acquisition for linear optimization.
\newblock In \emph{Uncertainty in artificial intelligence}, 2018.

\end{thebibliography}
\bibliographystyle{plainnat}

\clearpage
\appendix

\section{Omitted Proofs in the Main Text}
\label{app:proofs}
\begin{proof}[Proof of Theorem~\ref{thm:q0regret}]
Let $w^t_{i*} = \arg\min_{w} \mathbb E_{c_i^t | x_i^t}[c_i^t]^\dagger w$ and $w^t_{i} = \arg\min_{w} \hat c_i^{t^\dagger} w$.
The expected regret at round $t$ on individual $i$ is $\mathbb E[r_i^t]$, where 
\[
\begin{split}
r_i^t &= \mathbb E \left[ \mathbb E_{c_i^t|x_i^t}[c_i^t]^\dagger (w^t_{i} - w^t_{i*}) \right] \\
&\leq \mathbb E \left[  (\mathbb E_{c_i^t|x_i^t}[c_i^t] - \hat c_i^t)^\dagger (w^t_{i} - w^t_{i*}) \right] \\
&= O\left( \mathbb E \left[ \left\| \mathbb E_{c_i^t|x_i^t}[c_i^t] - \hat c_i^t\right\|_2 \right] \right)
\end{split}
\]
The first inequality above used the definition of $w_i^t$ and $w_{i*}^t$. The second step used Cauchy-Schwartz. Note that what remains to prove is simply an error bound on the OLS regression, which we prove as Lemma~\ref{lem:ols_error}. Using that result, we can conclude the total regret is
\[
\begin{split}
R_T &= \mathbb E[\sum_{t=1}^T \sum_{i=1}^n r_i^t]\\ 
&= O\left( \sum_{t=1}^T \sum_{i=1}^n \mathbb E \left[ \left\| \mathbb E_{c_i^t|x_i^t}[c_i^t] - \hat c_i^t\right\|_2 \right] \right)\\
&= O\left( \sum_{t=1}^T \sum_{i=1}^n \sqrt{\frac{dm}{nt}} \right) \\
&= O\left(\sqrt{ndmT} \right)    
\end{split}
\]
The last step (bounding $\sum_{t=1}^T t^{-1/2}$) is by an upper bound on the generalized harmonic numbers, which can be found in Theorem 3.2 (b) in the text by~\citet{apostolintroduction}.
\end{proof}

Recall that $\mathcal F$ is the class of all linear functions mapping $X$ to $C$ and $c = Fx + \epsilon$ where $F \in \mathcal F$ and $\epsilon \sim \mathcal N(0, \sigma^2 I_d)$.  Assume that $n > m$, that is, assume the number of data points we receive each round is greater than the number of features.
Let $F_k$ be the $k$-th row of $F$. Fix $k$, we have a linear regression problem $c_k = F_k^\dagger x + \epsilon_k$, where $\epsilon_k \sim \mathcal N(0, \sigma^2)$. At the $t$-th round, we have $nt$ data points and we need to predict on $n$ new data points. Let $X^t$ be the $n \times m$ matrix whose $i$-th row is $x^t_{i}$. Let $\tilde X^t$ be the $nt \times m$ matrix consisting of all the training data points. 
Suppose we fun an ordinary least wquares regression. Let $\hat F_k$ be the OLS estimate of $F_k$, and $\hat c_k = \hat F_k x$. 

\begin{lemma}
Suppose we use the ordinary least squares regression as the ML algorithm. The prediction error is 
\[
\mathbb E_{X, \epsilon} \left[ \left\| \mathbb E_{c_{i}^t | x_i^t}[c_{i}^t] - \hat c^t_{i} \right\|_2 \right] = O\left(\sqrt{\frac{dm}{nt}} \right),
\]
assuming either (1) $x \sim \mathcal N(0, \Lambda)$ follows a normal distribution, or (2) the eigenvalues of $\Sigma = \frac{\tilde X^{t^\dagger} \tilde X^t}{nt}$ are lower bounded by a positive number.
\label{lem:ols_error}
\end{lemma}

\begin{proof}
Consider the first case, since $x \sim \mathcal N(0, \Lambda)$,
we know $X^{T^\dagger} X^t \sim W(\Lambda, n)$, a Wishart distribution with $n$ degrees of freedom, and $\tilde X^{T^\dagger} \tilde X^t \sim W(\Lambda^{-1}, nt)$, an inverse Wishart distribution with $nt$ degrees of freedom. Thus, $\mathbb E_X[(X^{t^\dagger} X^t)^{-1}] = n\Lambda$ and $\mathbb E_X[(\tilde X^{t^\dagger} \tilde X^t)^{-1}] = \Lambda^{-1}/(nt-m-1)$. 
\[
\begin{split}
    &\mathbb E \left[\sum_{i=1}^n (\mathbb E_{c_{ik}^t | x_i^t}[c_{ik}^t] - \hat c^t_{ik})^2 \right] = \mathbb E \left[ ||X^t (F_k - \hat F_k)||_2^2 \right] \\
    &= \mathbb E \left[ (F_k - \hat F_k)^\dagger X^{t^\dagger} X^t (F_k - \hat F_k) \right] \\
    &= \mathbb E \left[ tr( (F_k - \hat F_k)^\dagger X^{t^\dagger} X^t (F_k - \hat F_k) )\right] \\
    &=  tr \left( \mathbb E \left[X^{t^\dagger} X^t\right] \mathbb E_{X}\left[ (F_k - \hat F_k) (F_k - \hat F_k)^\dagger \right] \right) \\
    &= \sigma^2  tr( \mathbb E\left[X^{t^\dagger} X^t\right] \mathbb E_{X}\left[ (\tilde X^{t^\dagger} \tilde X^t)^{-1} \right] )\\
    &= \sigma^2  tr\left(  \frac{n \Lambda \Lambda^{-1}}{nt-m-1}  \right) = \frac{nm\sigma^2 }{nt-m-1}
\end{split}
\]
The above derivation has appeared in previous literature, e.g. the work by~\citet{rosset2020fixed}. The result holds for all $k$, we get 
\[
\mathbb E_{X, \epsilon} \left[ \left\| \mathbb E_{c_{i}^t | x_i^t}[c_{i}^t] - \hat c^t_{i} \right\|_2^2 \right]  = \frac{md\sigma^2 }{nt-m-1}.
\]
That is,
\[
\mathbb E_{X, \epsilon} \left[ \left\| \mathbb E_{c_{i}^t | x_i^t}[c_{i}^t] - \hat c^t_{i} \right\|_2 \right] = O\left(\sqrt{\frac{dm}{nt}} \right).
\]

In the second case, suppose the eigenvalues of $\Sigma = \frac{\tilde X^{t^\dagger} \tilde X^t}{nt}$ are lower bounded by a constant $K_\Sigma > 0$.
\[
\begin{split}
    &\mathbb E_{X, \epsilon} \left[ \left|\mathbb E_{c_{ik}^t | x_i^t}[c_{ik}^t] - c^t_{ik} \right| \right] =  \mathbb E_{X, \epsilon} \left[  \left| F_k^\dagger  x^t_{ik} - \hat F_k^\dagger x^t_{ik} \right| \right]\\
    &\leq \mathbb E_{X} \left[ \left|F_k^\dagger  x^t_{ik} - \hat F_k^\dagger x^t_{ik} \right| \right] 
    \leq \mathbb E_{X} \left[ \left\| F_k- \hat F_k \right\|_2  \left\| x^t_{ik} \right\|_2 \right] \\
    &\leq K_X \mathbb E_{X} \left[ \left\| F_k- \hat F_k \right\|_2^2  \right]^{1/2} 
    = K_X \mathbb E_X \left[ tr(\sigma^2 (\tilde X^{t^\dagger} \tilde X^t)^{-1} ) \right]^{1/2} \\
    &=  \frac{\sigma K_X}{\sqrt{nt}} \mathbb E_X \left[ tr \left(\Sigma^{-1} \right) \right]^{1/2}
\end{split}
\]
Then the prediction error can be bounded by
\[
    \mathbb E_{X, \epsilon} \left[ \left|\mathbb E_{c_{ik}^t | x_i^t}[c_{ik}^t] - \hat c^t_{ik} \right| \right] \leq  O\left( \sqrt{\frac{m}{nt}} \right)
\]
This holds for all $k$. Thus, we have
\[
    \mathbb E_{X, \epsilon} \left[ \left\|\mathbb E_{c_{i}^t | x_i^t}[c_{i}^t] - \hat c^t_{i} \right\|_2 \right] \leq  O\left( \sqrt{\frac{md}{nt}} \right) 
\]
\end{proof}

\begin{proof}[Proof of Lemma~\ref{lem:regret}]
Let $w^t_{i*} = \arg\min_{w} (\mathbb E_{c_i^t | x_i^t}[c_i^t] + \mu)^\dagger w$. $w^t_{i*}$ is the optimal action for individual $i$ at time $t$, and is the benchmark in our regret computation.

Fix $i$, fix $t$. Let $\tilde \nu = \arg\min_{\nu \in B^t_i} (\hat c^t_i + \nu)^\dagger w^t_i$.
Because of Line~\ref{algline:merge}, we have
\[
\begin{split}
&(\hat c^t_i + \tilde \nu)^\dagger w^t_i = \min_{\nu\in B^t_i, w\in W} (\hat c^t_i + \nu)^\dagger w \\
&\leq (\mathbb E_{c_i^t | x_i^t}[c_i^t] + \mu)^\dagger w^t_{i*} + (\hat c^t_i)^\dagger w^t_{i*} - \mathbb E_{c_i^t | x_i^t}[c_i^t]^\dagger w^t_{i*}.
\end{split}
\]
The inequality above used the fact that $\mu \in B_i^t$, by Lemma~\ref{lem:confidenceball}.
Thus, we get the per-round regret
\[
\begin{split}
&(\mathbb E_{c_i^t | x_i^t}[c_i^t] + \mu)^\dagger (w^t_i - w^t_{i*}) \\
&\leq (\mathbb E_{c_i^t | x_i^t}[c_i^t] + \mu)^\dagger w^t_i - (\hat c^t_i + \tilde \nu)^\dagger w^t_i + (\hat c^t_i)^\dagger w^t_{i*} \\
&- \mathbb E_{c_i^t | x_i^t}[c_i^t]^\dagger w^t_{i*}\\
&= (\mathbb E_{c_i^t | x_i^t}[c_i^t] - \hat c^t_i)^\dagger (w^t_i - w^t_{i*}) + (\mu - \tilde \nu)^\dagger w^t_i
\end{split}
\]
We can view the second term is the per-round regret for the bandit part. By Theorem 6 in~\citep{dani2008stochastic}, we have
\[
\sum_{t=1}^T ((\mu - \tilde \nu)^\dagger w^t_i)^2 \leq 8m \beta^T \log T
\]
Using the Cauchy-Schwarz, we get 
\[
\sum_{t=1}^T (\mu - \tilde \nu)^\dagger w^t_i \leq \sqrt{8m T \beta^T \log T}
\]
Thus, the regret of Algorithm~\ref{alg:p2pofu} is
\[
\begin{split}
& \mathbb E \left[ \sum_{t=1}^T \sum_{i=1}^n (\mathbb E_{c_i^t | x_i^t}[c_i^t] + \mu)^\dagger (w^t_i - w^t_{i*}) \right] \\
&\leq \mathbb E \left[ \sum_{t=1}^T \sum_{i=1}^n (\mathbb E_{c_i^t | x_i^t}[c_i^t] - \hat c^t_i)^\dagger (w^t_i - w^t_{i*}) \right] \\
&+ n\sqrt{8mT \beta_T \log T} \\
& = O\left(  \sum_{t=1}^T \sum_{i=1}^n \mathbb E \left[ \left\| \mathbb E_{c_i^t | x_i^t}[c_i^t] - \hat c^t_i \right\|_2 \right] + n\sqrt{8mT \beta_T \log T} \right) \\
\end{split}
\]
The last step used Cauchy-Schwartz and the bounded action space assumption.
\end{proof}

\begin{proof}[Proof of Theorem~\ref{thm:regret_linear}]
Combine Lemma~\ref{lem:regret} and Lemma~\ref{lem:ols_error}.
\end{proof}

\begin{proof}[Proof of Lemma~\ref{lem:regret_cf}]
In Algorithm~\ref{alg:p2pofucf}, at each round $1,2,\dots, \tilde T$ in the exploration phase, each action is sequentially taken on some examples. This means by the end of step $\tilde T$, we have $\tilde n = n\tilde T/|W|$ data points for training the predictor for each $w_i$. Although in practice one can keep updating (learning) the predictor during the exploitation phase, in the following theoretical analysis it suffices to ignore this additional learning effect.
We assume $\tilde n$ is an integer, but this is not an issue in the general case.

Let us first analyze the regret in the exploitation phase.

Let $w^t_{i*} = \arg\min_{w} (\mathbb E_{c_i^t | x_i^t, w} [c^t_i(w)] + \mu)^\dagger w$. $w^t_{i*}$ is the optimal action for individual $i$ at time $t$, and is the benchmark in our regret computation.

Fix $i$, fix $t$. Let $\tilde \nu = \arg\min_{\nu \in B^t_i} (\hat c^t_i(w^t_i) + \nu)^\dagger w^t_i$.
Because of Line~\ref{algline:merge_cf}, we have
\[
\begin{split}
&(\hat c^t_i(w^t_i) + \tilde \nu)^\dagger w^t_i = \min_{\nu\in B^t_i, w\in W} (\hat c^t_i(w) + \nu)^\dagger w \\
&\leq (\mathbb E_{c_i^t | x_i^t, w^t_{i*}} [c^t_i(w^t_{i*})] + \mu)^\dagger w^t_{i*} + (\hat c^t_i(w^t_{i*}))^\dagger w^t_{i*}\\
&-  \mathbb E_{c_i^t | x_i^t, w^t_{i*}} [c^t_i(w^t_{i*})]^\dagger w^t_{i*}.
\end{split}
\]
The inequality above used the fact that $\mu \in B_i^t$, by Lemma~\ref{lem:confidenceball}.
Thus, we get the per-round regret
\[
\begin{split}
&(\mathbb E_{c_i^t | x_i^t, w_i^t} [c^t_i(w_i^t)] + \mu)^\dagger w^t_i - (\mathbb E_{c_i^t | x_i^t, w^t_{i*}} [c^t_i(w^t_{i*})] + \mu)^\dagger w^t_{i*}  \\
&\leq (\mathbb E_{c_i^t | x_i^t, w_i^t} [c^t_i(w_i^t)] + \mu)^\dagger w^t_i  - (\hat c^t_i(w^t_i) + \tilde \nu)^\dagger w^t_i  \\
&+ (\hat c^t_i(w^t_{i*}))^\dagger w^t_{i*} - \mathbb E_{c_i^t | x_i^t, w^t_{i*}} [c^t_i(w^t_{i*})]^\dagger w^t_{i*}\\
&= (\mathbb E_{c_i^t | x_i^t, w_i^t} [c^t_i(w_i^t)] - \hat c^t_i(w^t_i))^\dagger w^t_i \\
&+ (\hat c^t_i(w^t_{i*}) -\mathbb E_{c_i^t | x_i^t, w^t_{i*}} [c^t_i(w^t_{i*})])^\dagger w^t_{i*} + (\mu - \tilde \nu)^\dagger w^t_i
\end{split}
\]
We can view the third term as the per-round regret for the bandit part. By Theorem 6 in~\citep{dani2008stochastic}, we have
\[
\sum_{t=1}^T ((\mu - \tilde \nu)^\dagger w^t_i)^2 \leq 8m \beta^T \log T
\]
Using the Cauchy-Schwarz, we get 
\[
\sum_{t=1}^T (\mu - \tilde \nu)^\dagger w^t_i \leq \sqrt{8m T \beta^T \log T}
\]
Thus, the regret of Algorithm~\ref{alg:p2pofucf} is upper bounded by
\[
\begin{split}
& \mathbb E \left[ \sum_{t=1}^T \sum_{i=1}^n (\mathbb E_{c_i^t | x_i^t, w_i^t} [c^t_i(w_i^t)] + \mu)^\dagger w^t_i - (\mathbb E_{c_i^t | x_i^t, w^t_{i*}} [c^t_i(w^t_{i*})] + \mu)^\dagger w^t_{i*} \right]\\
& \leq K n \tilde T + \mathbb E \Bigg[ \sum_{t=\tilde T + 1}^T \sum_{i=1}^n (\mathbb E_{c_i^t | x_i^t, w_i^t} [c^t_i(w_i^t)] - \hat c^t_i(w^t_i))^\dagger w^t_i\\
&+ (\hat c^t_i(w^t_{i*}) -\mathbb E_{c_i^t | x_i^t, w^t_{i*}} [c^t_i(w^t_{i*})])^\dagger w^t_{i*} \Bigg] + n\sqrt{8mT \beta_T \log T} \\
& = O\Bigg( n\tilde T + n\sqrt{8mT \beta_T \log T} \\
&+ \sum_{t=\tilde T + 1}^T \sum_{i=1}^n \mathbb E \left[ \left\| \mathbb E_{c_i^t | x_i^t, w_i^t} [c^t_i(w_i^t)] - \hat c^t_i(w^t_i) \right\|_2 \right]  \Bigg)
\end{split}
\]
The last step used Cauchy-Schwartz, symmetry, and the bounded action space assumption.
\end{proof}

\begin{proof}[Proof of Theorem~\ref{thm:regret_linear_cf}]
Again, we prove by bounding the linear regret prediction error. The proof follows identically as Theorem~\ref{thm:regret_linear}. We get, $ \forall t > \tilde T, \forall w, \forall i$,
\[
\mathbb E_{X, \epsilon} \left[ \left\| \mathbb E_{c_i^t | x_i^t, w_i^t} [c^t_i(w_i^t)] - \hat c^t_i(w^t_i) \right\|_2 \right] = O\left(\sqrt{\frac{dm |W|}{n\tilde T}} \right).
\]
Using Lemma~\ref{lem:regret_cf}, and taking $\tilde T = T^{2/3} (d|W|)^{1/3}$, we get
\[
\begin{split}
    & O\left( n\tilde T + n\sqrt{8mT \beta_T \log T} + \sum_{t=\tilde T + 1}^T \sum_{i=1}^n \sqrt{\frac{dm |W|}{n\tilde T}}  \right)\\
    &= O\left( n\tilde T + n\sqrt{8mT \beta_T \log T} + T  \sqrt{\frac{ndm |W|}{\tilde T}}  \right)\\
    &= \tilde O\left( (d |W|)^{1/3} m^{1/2} n T^{2/3}  \right)
\end{split}
\]
\end{proof}

\begin{proof}[Proof of Theorem~\ref{thm:regret_linear_cf_cont}]
Using Lemma~\ref{lem:ols_error}, we know that at the beginning of the exploitation phase, the prediction error is 
\[
\mathbb E_{X, \epsilon} \left[ \left\| \mathbb E_{c_i^t | x_i^t, w_i^t} [c^t_i(w_i^t)] - \hat c^t_i(w^t_i) \right\|_2 \right] = O\left(\sqrt{\frac{(m+d)d}{n\tilde T}} \right),
\]
$\forall t > \tilde T, \forall w, \forall i$.
Thus, using Lemma~\ref{lem:regret_cf}, we know the regret is $\tilde O\left( m^{1/3} d^{2/3} n T^{2/3} \right)$, when we take $\tilde T = m^{1/3} d^{2/3} T^{2/3}$. 
\end{proof}

\section{Omitted Algorithm}
\label{app:algo}
\begin{algorithm}[!h]
\KwInitialize{Find a barycentric spanner $b_1,\dots, b_n$ for $W$\\
Set $A^1_i = \sum_{j=1}^d b_j b_j^\dagger$ and $\hat \mu^1_i = 0$ for all $i = 1,2,\dots, n$\\
}
Receive initial dataset $\mathcal D = \{(x_i^{0}, c_i^{0}; w^{0}_i)_{i=1,\dots,n}\}$ from distribution $D$ on $(X, C)$.\\
\tcp{Uniform exploration phase}
\For{$t=1,2,\dots, \tilde T$  }{
		\For{$i=1,2,\dots, n$}{
		Given feature sample $x_i^t$, choose intervention $w^t_i = w_{nt+i \mod |W|}$ where $W = \{w_1,\dots, w_{|W|}\}$ is considered as an ordered set. \\
		Receive label $c_i^t \sim D(w_i^t)_{c|x_i^t}$. Add $(x_i^t, c_i^t; w_i^t)$ to the dataset $\mathcal D$. \\
		Get cost $u^t_i = u(x^t_i, c_i^t, w^t_i) = (c^t_i(w^t_i))^\dagger w_i^t + \mu^\dagger w^t_i + \eta_i$, where $\eta_i \sim N(0, \sigma^2)$. In particular, let $u^t_{oi}$ be the first term and let $u^t_{bi}$ be the sum of the second and third term.\\ 
		Update $A^{t+1}_i = A^t_i + w^t_i (w^t_i)^\dagger$ \\ 
		Update $\hat \mu^{t+1}_i = (A^{t+1}_i)^{-1} \sum_{\tau=1}^t u^t_{bi} w^t_i$
		}
}
\tcp{UCB exploitation phase}
	\For{$t = \tilde T+1, \dots, T$}{
		For each $w$, using all the available data $\mathcal D$ that were collected under $w$, train ML prediction model $f^t_w: X \to C$.\\ 
		Given $n$ feature samples $\{x_i^t\} \sim D_x$, get predictions $\hat c_i^t(w) = f_t(x^t_i)$, for each $w$.\\
		Set confidencec ball radius $\beta^t = \max\left( 128 d\log t \log(n t^2/\gamma), \left( \frac{8}{3} \log \left(\frac{n t^2}{\gamma} \right) \right)^2 \right)$\\
		\For{$i=1,2,\dots, n$}{
	        Set confidence ball $B_i^t = \{\nu: ||\nu - \hat \mu^t_i||_{2, A^t_i} \leq \sqrt{\beta^t} \}$.\\
		Solve optimization problem $w^t_i = \arg\min_{w\in W} \min_{\nu \in B^t_i} (\hat c^t_i(w) + \nu)^\dagger w$. Choose intervention $w^t_i$. \\ \label{algline:merge_cf}
		Receive label $c_i^t \sim D(w_i^t)_{c|x_i^t}$. Add $(x_i^t, c_i^t; w_i^t)$ to the dataset $\mathcal D$. \\
		Get cost $u^t_i = u(x^t_i, c_i^t, w^t_i) = (c^t_i(w^t_i))^\dagger w_i^t + \mu^\dagger w^t_i + \eta_i$, where $\eta_i \sim N(0, \sigma^2)$. In particular, let $u^t_{oi}$ be the first term and let $u^t_{bi}$ be the sum of the second and third term.\\ 
		Update $A^{t+1}_i = A^t_i + w^t_i (w^t_i)^\dagger$ \\ 
		Update $\hat \mu^{t+1}_i = (A^{t+1}_i)^{-1} \sum_{\tau=1}^t u^t_{bi} w^t_i$ 
		}
	}
	\caption{\textsc{PROOF with action-specific label distribution }}\label{alg:p2pofucf}
\end{algorithm}

\section{Regret Bounds Using Sample Complexity Characterization}
\label{app:rad}
Let us first introduce the multivariate Rademacher complexity and its associated generalization bounds, which were introduced by~\cite{bertsimas2020predictive}.

\begin{definition}
Given a sample $S_n = \{s_1,\dots, s_n \}$, the empirical multivariate Rademacher complexity of a class of functions $\mathcal F$ taking values in $\mathbb R^d$ is defined as
\[
\hat{\mathfrak{R}}_n(\mathcal F; S_n) = \mathbb E_{\sigma} \left[ \sup_{g \in \mathcal F}  \frac{1}{n} \sum_{i=1}^n \sum_{k=1}^d \sigma_{ik} g_k(s_i) \right]
\]
where $\sigma_{ik}$'s are independent Rademacher random variables.
The multivariate Rademacher complexity is defined as $\mathfrak{R}_n(\mathcal F) = \mathbb E_{S_n} [\hat{\mathfrak{R}}_n(\mathcal F; S_n)]$
\end{definition}
\begin{theorem}~\citep{bertsimas2020predictive}
Suppose function $c(z; y)$ is bounded and equi-Lipschitz in $z$:
\[
\sup_{z \in Z, y \in Y} c(z; y) \leq \bar c,\,\text{and} \sup_{z \neq z' \in Z, y \in Y} \frac{c(z;y) - c(z';y)}{||z - z'||_\infty} \leq L < \infty
\]
For any $\delta > 0$, each of the following events occurs with probability at least $1-\delta$,
\begin{align*}
&\mathbb E[c(z(X);Y )] \\
&\leq \frac{1}{n} \sum_{i=1}^n c(z(x^i); y^i) + L \mathfrak{R}_n(\mathcal F) + \bar c \sqrt{\frac{\log(1/\delta)}{2n}}\\
&\mathbb E[c(z(X);Y )] \\
&\leq \frac{1}{n} \sum_{i=1}^n c(z(x^i); y^i) + L \hat{\mathfrak{R}}_n(\mathcal F; S_n) + 3\bar c \sqrt{\frac{\log(2/\delta)}{2n}}.
\end{align*} \label{thm:rademacher}
\end{theorem}

When the function $c(z; y)$ is nonnegative, we have
\begin{equation}
\begin{split}
&\mathbb E[c(z(X);Y )] \label{eqn:exp_rad}\\
&\leq \frac{1}{1-\delta} \mathbb E \left[ \frac{1}{n} \sum_{i=1}^n c(z(x^i); y^i) \right] + L \mathfrak{R}_n(\mathcal F) + \bar c \sqrt{\frac{\log(1/\delta)}{2n}}
\end{split}
\end{equation}

Before we proceed, we make the following assumption.

\begin{assumption}
With $n$ training data points $x_1,\dots x_n$, the learning algorithm learns a predictor $\hat f$ such that $\mathbb E \left[ \sum_{i=1}^n   \left\| f(x_i) - \hat f(x_i) \right\|_2^2  \right]$
is constant with respect to $n$.
\label{asp:constantER}
\end{assumption}

Although this assumption might appear somewhat unintuitive, it is actually satisfied when, for example, $f \in \mathcal F$ comes from the class of all linear functions, ordinary least squares regression used as the learning algorithm satisfies this assumption. In that case, we have $\mathbb E \left[ \sum_{i=1}^n   \left\| f(x_i) - \hat f(x_i) \right\|_2^2  \right] = O(md)$.

\begin{theorem}
Suppose we use any learning algorithm that satisfies Assumption~\ref{asp:constantER}, including but not limited to OLS regression. The regret of Algorithm~\ref{alg:p2pofu} is $\tilde O\left(md \sqrt{nT} \right)$, with probability $1 - \delta$.
\label{thm:regret_linear_rad}
\end{theorem}
\begin{proof}
First, let's compute the Rademacher complexity of the linear hypothesis class, for completeness and for our specific setting.
Let $F_k$ be the $k$-th row of matrix $F$. We have
\[
\begin{split}
    &\hat{\mathfrak{R}}_n(\mathcal F; X_n)
    = \mathbb E_{\sigma} \left[ \sup_{F \in \mathcal F}  \frac{1}{n} \sum_{i=1}^n \sum_{k=1}^d \sigma_{ik} F_k^\dagger x_i \right] \\
    &= \mathbb E_{\sigma} \left[ \sup_{F \in \mathcal F}  \frac{1}{n} \sum_{k=1}^d  F_k^\dagger \left(\sum_{i=1}^n  \sigma_{ik}  x_i \right) \right] \\
    &\leq \mathbb E_{\sigma} \left[ \sup_{F \in \mathcal F}  \frac{1}{n} \sum_{k=1}^d  ||F_k||_1 \left\| \sum_{i=1}^n  \sigma_{ik}  x_i \right\|_2 \right] \quad \text{(Cauchy-Schwartz)}\\
    &\leq \mathbb E_{\sigma} \left[ \sup_{F \in \mathcal F}  \frac{1}{n} \sum_{k=1}^d  ||F||_\infty \left\| \sum_{i=1}^n  \sigma_{ik}  x_i \right\|_2 \right]\\
    &\leq \sum_{k=1}^d  m K_F \mathbb E_{\sigma} \left[ \left\| \frac{1}{n}  \sum_{i=1}^n  \sigma_{ik}  x_i \right\|_2 \right] \\
    &\leq \sum_{k=1}^d  m K_F \left(\mathbb E_{\sigma} \left[ \left\| \frac{1}{n}  \sum_{i=1}^n  \sigma_{ik}  x_i \right\|_2^2 \right] \right)^{1/2} \qquad \text{(Jensen)}\\
    &\leq \sum_{k=1}^d  m K_F \left(\mathbb E_{\sigma} \left[  \frac{1}{n^2}  \sum_{i=1}^n  ||x_i||_2^2 + \frac{2}{n^2}  \sum_{i<j} \sigma_{ik} \sigma_{jk} x_i^\dagger x_j  \right] \right)^{1/2}\\
    &= \sum_{k=1}^d  m K_F \left(\mathbb E_{\sigma} \left[  \frac{1}{n^2}  \sum_{i=1}^n  ||x_i||_2^2   \right] \right)^{1/2} =  \frac{d m K_F K_X}{\sqrt{n}}  \\
\end{split}
\]
Using Equation~\ref{eqn:exp_rad}, we have
\[
\begin{split}
 &\mathbb E_{X, \epsilon} \left[ \left\|\mathbb E_{c_{i}^t | x_i^t}[c_{i}^t] - \hat c^t_{i} \right\|_2 \right]\\
 &\leq  \frac{1}{(1-\delta)nt} \mathbb E \left[ \sum_{i=1}^n \sum_{\tau=1}^t  \left\| f(x_i^\tau) - \hat f(x_i^\tau) \right\|_2 \right] \\
 &+ L \mathfrak{R}_{nt}(\mathcal F) + \bar c \sqrt{\frac{\log(T/\delta)}{2nt}} \\
 &\leq \frac{1}{(1-\delta)nt} \mathbb E \left[ \sqrt{ nt \sum_{i=1}^n \sum_{\tau=1}^t  \left\| f(x_i^\tau) - \hat f(x_i^\tau) \right\|_2^2 } \right] \\
 &+ L \mathfrak{R}_{nt}(\mathcal F) + \bar c \sqrt{\frac{\log(T/\delta)}{2nt}} \\
 &\leq \frac{1}{(1-\delta)\sqrt{nt}} \sqrt{ \mathbb E \left[ \sum_{i=1}^n \sum_{\tau=1}^t  \left\| f(x_i^\tau) - \hat f(x_i^\tau) \right\|_2^2  \right] } \\
 &+ L \mathfrak{R}_{nt}(\mathcal F) + \bar c \sqrt{\frac{\log(T/\delta)}{2nt}}
 \end{split}
\]
By Assumption~\ref{asp:constantER}, the term under square root is constant w.r.t. $nt$, under regularity conditions. Thus, we have
\[
\mathbb E_{X, \epsilon} \left[ \left\|\mathbb E_{c_{i}^t | x_i^t}[c_{i}^t] - \hat c^t_{i} \right\|_2 \right] = \tilde O\left( md (nt)^{-1/2} \right)
\]
The rest of the proof follows from Lemma~\ref{lem:regret}. 
\end{proof}

This result is almost identical as Theorem~\ref{thm:regret_linear}. However, the intent to work with Rademacher complexity is that we hope to at least get some bound when we move beyond the linear regression scenario. Let us consider a feed-forward neural network with ReLU activation. There are existing results which shows that the Rademacher complexity is $O(1/\sqrt{n})$~\citep{golowich2018size}. Thus, if we accept Assumption~\ref{asp:constantER}, we would have the following result.

\begin{theorem}
Supose the learning problem is fitting a neural network and we use any learning algorithm that satisfies Assumption~\ref{asp:constantER}. The regret of Algorithm~\ref{alg:p2pofu} is $\tilde O(n T^{1/2})$, with probability $1-\delta-\lambda$, ignoring the dependency on $d$ and $m$.
\end{theorem}

Finally, we extend the previous results to the more general case where interventions affect the label distribution. Please refer to the setting described in Section~\ref{sec:cf} and Algorithm~\ref{alg:p2pofucf} in Appendix~\ref{app:algo}. 
\begin{theorem}
Suppose there are finitely many actions. Assuming we use any learning algorithm that satisfies Assumption~\ref{asp:constantER}, including but not limited to OLS regression, the regret of Algorithm~\ref{alg:p2pofu} is $\tilde O\left(|W|^{1/3} (md)^{2/3} n T^{2/3} \right)$, with probability $1 - \delta$.
\label{thm:regret_linear_rad_cf}
\end{theorem}
\begin{proof}
After the exploration phase, we have had $\tilde n = n\tilde T/|W|$ data points for training the predictor for each $w_i$. 
Similar to our approach in Theorem~\ref{thm:regret_linear_rad}, we have
\[
\begin{split}
&\mathbb E \left[ \left\| \mathbb E_{c_i^t | x_i^t, w_i^t} [c^t_i(w_i^t)] - \hat c^t_i(w^t_i) \right\|_2 \right] \\
&\leq  \frac{1}{(1-\delta)\tilde n} \mathbb E \left[ \sum_{i=1}^{\tilde n} \left\| f(x_i) - \hat f(x_i) \right\|_2^2 \right] \\
&+ L \mathfrak{R}_{\tilde n}(\mathcal F) + \bar c \sqrt{\frac{\log(T/\delta)}{2\tilde n}} \\
 &= \tilde O\left(|W|^{1/2} md (n \tilde T)^{-1/2}  \right),\qquad \forall t > \tilde T
 \end{split}
\]
Thus, the regret is
\[
\begin{split}
&O\Bigg( n\tilde T + n\sqrt{8mT \beta_T \log T} \\
&+ \sum_{t=\tilde T + 1}^T \sum_{i=1}^n \mathbb E \left[ \left\| \mathbb E_{c_i^t | x_i^t, w_i^t} [c^t_i(w_i^t)] - \hat c^t_i(w^t_i) \right\|_2 \right]  \Bigg)\\
&= \tilde O\left( n\tilde T + n\sqrt{8mT \beta_T \log T} + nT |W|^{1/2} dm(n \tilde T)^{-1/2}  \right)\\
&= \tilde O\left(  |W|^{1/3} (md)^{2/3} n T^{2/3} \right)
\end{split}
\]
where we let $\tilde T = T^{2/3} |W|^{1/3} (md)^{2/3}$. 
\end{proof}

\section{Details of the Food Rescue Experiment}
\label{app:fr}

In this section, we provide additional information about the ML recommender system component of the food rescue experiment in Section~\ref{sec:frexperiment}. The data description and feature engineering in Appendix~\ref{app:frdata} is adapted from~\citep{shi2021recommender} to serve our purpose. The design and training of the recommender system in Appendix~\ref{app:frmodel} and Appendix~\ref{app:frtraining} are novel of this work.

\subsection{Data}
\label{app:frdata}
To develop a recommender system, we need both positive and negative labeled examples. A positive example means that a particular volunteer (item) claims a particular rescue (user); a negative example means the volunteer does not claim the rescue. In this section, we detail our data acquisition, labeling, and feature engineering process.

\paragraph{Positive Labels}
We obtained the rescue database from our partner organization, covering the period from March 2018 to March 2020. The database keeps the log of each rescue. For most rescues, the database logs its timestamps from being drafted by the dispatcher, to being published on the mobile app, to being claimed and completed by a volunteer. We take the rescue plus the volunteer who claimed it as a positive data point. 

\paragraph{Negative Labels}
\label{sec:neglabels}
A negative example means that a particular volunteer did not claim a particular rescue. Since almost all rescues have only one volunteer who claimed the rescue, obviously most of our data points will have negative labels. However, not all of these negative data points are necessarily true, because perhaps a volunteer would have claimed some rescue if someone else had not claimed it 10 minutes in advance. Thus, we use the following ways to construct a selected negative dataset. First, in the time period covered by our database, our partner used a mobile app push notification scheme which notifies volunteers within 5 miles when the rescue is first available and then notifies all volunteers 15 minutes later if the rescue has not been claimed. Thus, if a rescue is claimed within 15 minutes, we only treat the volunteers who were within 5 miles and did not opt out of push notifications as negative examples. 

We also incorporate another data source to strengthen our negative sampling. In addition to mobile app notifications, the dispatcher at our partner organization also manually call some regular volunteers to ask for help with a specific rescue. This usually happens when some rescue has been available for over an hour yet nobody has claimed it. We obtained the call history, from which we identify the volunteers they reached out to within the time frame of each rescue. If these volunteers did not claim the rescues in the end, we treat them as negative examples. Compared to the negative examples derived from push notifications, we have more confidence in this set of negative examples, since declining on a phone call is a stronger indicator than ignoring a push notification.

\paragraph{Feature Engineering}

\begin{figure}[t]
\centering
    \subfloat[Distribution of donor organizations. Darker colors mean more frequent donations. We plot the donor locations with random perturbations. ]{\includegraphics[width=0.47\columnwidth]{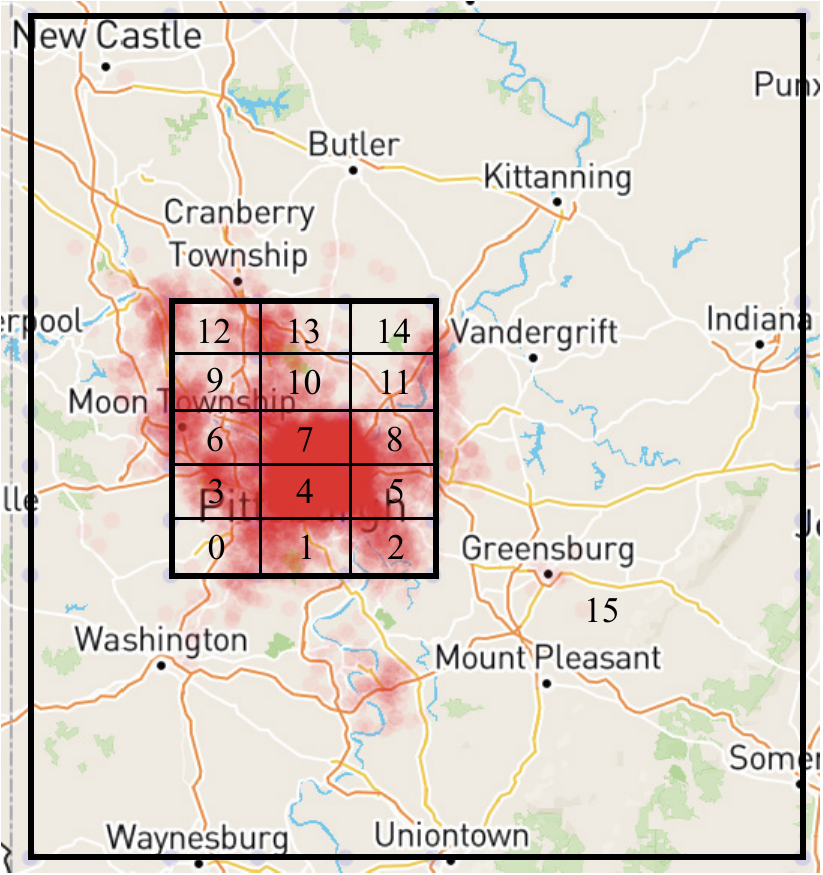}\label{fig:donor_grid}} \quad
    \subfloat[Density of recipient organizations. Darker colors mean more recipient organizations in the grid.]{\includegraphics[width=0.47\columnwidth]{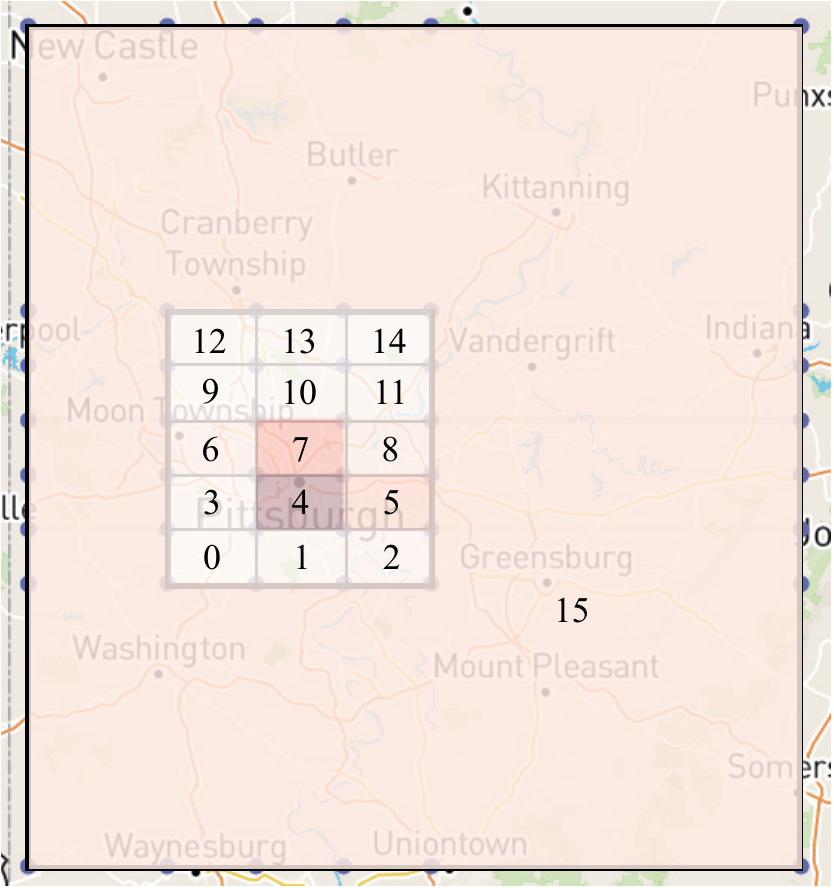}\label{fig:recipient_grid}}
    \caption{We divide the area of interest into 16 grid cells, with cells 0--14 covering the downtown and its neighborhoods, and cell 15 containing the rest of the region.}
    \label{fig:grid}
\end{figure}

Based on our collaboration with our partner, we carefully identify a selected set of useful features that are relevant in the food rescue operation.

First, the experience of food rescue dispatcher indicates that if a volunteer has completed a rescue at or near a donor or recipient, they are more likely to do a rescue trip again in the neighborhood. As shown in Figure~\ref{fig:grid}, we divide the region of interest into 16 cells. We evenly divide a central rectangular region into a $3 \times 5$ grid, and label them grid cells $0$ through $14$. Then, we label the entire map outside the rectangular region cell $15$. The rationale is that in the outer suburbs there are fewer donors, recipients, and volunteers, and furthermore volunteers who live in suburbs are more willing to do long-distance, i.e. inter-cell, rescue than volunteers in downtown. For each rescue trip and each volunteer, we calculate the number of rescues the volunteer has done in the rescue donor's cell, in the rescue recipient's cell, and across all cells. We also tried to include as features the volunteer's historical rescues in each cell, not just the donor's and recipient's cell. However, they did not contribute any predictive power and thus we leave them out of the final model.

Closely related to this is the distance between the volunteer and the donor. It is unlikely that a volunteer would drive 30 miles to pick up a donation. We measure the distance using the straight line distance based on geographic coordinates. Although the actual traveling distance might be a better indicator, we observe that the straight line distance already serves our purpose. 

\begin{figure}[t]
\centering
    \includegraphics[width=0.8\columnwidth]{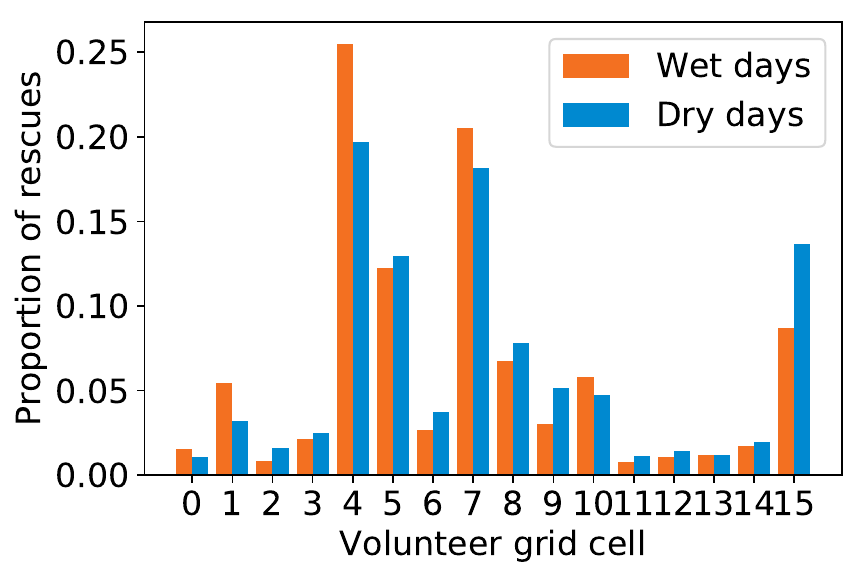}
    \caption{Histograms of rescues under wet and dry weather, based on the location of the volunteer who claimed the rescue.}
    \label{fig:prcp_grid}
\end{figure}

Aside from the geographical information, the length of time between volunteer's registration on the platform and the rescue is also an important factor, as suggested by our partner. Immediately after registration, the volunteer is eager to claim a rescue to get a feel of the food rescue experience. Thus, we include this feature in our prediction model. 

Weather information is also an important factor in the prediction. Presumably rainy and snowy days would see a lower volunteer activity in general. However, the impact of inclement weather would fall disproportionately on volunteers who do not have a car or live in suburban areas. We use the Climate Data Online (CDO) service provided by the National Oceanic and Atmospheric Administration to access the weather information.\footnote{\url{https://www.ncdc.noaa.gov/cdo-web/}} The CDO dataset contains weather information at the discretization level of days and weather station. There are multiple weather stations in the area and for each rescue we select the data for the date of rescue and the station that is closest to the donor organization. As shown in Figure~\ref{fig:prcp_grid}, on wet days, relatively more volunteers who claim the rescue reside in downtown (cell 4 and 7). Whereas on dry days, a lot more volunteers who live in the outer suburbs (cell 15) are active. In fact, we also saw a significant difference in the average distance between volunteer and donor for dry days (5.94 miles) and rainy days (5.22 miles), with a t-test p-value $3\times 10^{-8}$.

We also explored a number of other features but did not incorporate them into our final model. These features include the rescue's time of day and day of week, the volunteer's availability, whether the volunteer uploaded an avatar to their profile or not, whether the volunteer is located in the same grid as the donor or recipient, and so on. Although these are intuitive factors, we did not find them improve the predictive power of our model and hence left them out.

\subsection{Recommender System Model}
\label{app:frmodel}
We build our recommender system using a neural network.
We show the neural network architecture in Table~\ref{tab:architecture}. The input to the neural network is the feature vector of a rescue-volunteer pair. The feature vector passes through three dense layers. Each layer is followed by a ReLU activation function, except for the last layer where we output a single number which is then converted to a number between 0 and 1 by the logistic function. This output represents the likelihood that this volunteer will claim this rescue trip. We use the cross entropy loss to train the neural network. At prediction time, for a given rescue, we pass the feature vectors of the rescue-volunteer pairs for all volunteers on a fixed rescue through the network and obtain a likelihood estimate for each volunteer.

\begin{table}[t]
    \centering
    \caption{Neural network architecture}
    \label{tab:architecture}
    \begin{tabular}{ccc} \hline
        Layer & Operation & Hidden Units \\ \hline
        1 & Dense (ReLU) & 192 \\
        2 & Dense (ReLU) & 512 \\
        3 & Dense (Logistic) & 16\\ \hline
    \end{tabular}
\end{table}

\subsection{Training}
\label{app:frtraining}
We performed all the experiments in this paper on an Intel Core i5-7600K CPU and 32GB RAM.

We use the data from March 2018 to October 2019 for feature preparation. Recall that some of the features we use are related to the volunteer's historical number of rescues. We use the data from this period to generate such features. Then, we use the 556 rescues from November 2019 to March 2020 for learning and prediction in the actual experiment. In this way we avoid the potential data leakage.

In these 556 rescues, we select the first 300 of them to be the initial dataset for the bandit data-driven optimization (refer to Line~\ref{algline:dataset} in Procedure~\ref{alg:procedure}, Section~\ref{sec:model}). From the remaining 256 rescues, we randomly sample and set aside 150 rescues as the validation dataset. Finally we take the 50 earliest rescues from the remaining 106 rescues to run the PROOF algorithm for 50 iterations, each iteration corresponding to one rescue. At time step $t$, our training set consists of the 300 rescues in the initial dataset and all the rescues we have seen from time step 1 up to time step $t-1$. When training the recommender system at each time step, we use the Adam optimizer with learning rate $1 \times 10^{-3}$. We stop the training when the 3-episode moving average loss on the validation set stops decreasing. In the following paragraph, we discuss our way to address a key challenge in the training dataset in more detail.

\paragraph{Negative Sampling}
\label{sec:negsampling}
As mentioned earlier, there is an extremely high label imbalance in our dataset. Each rescue typically has only one volunteer who claimed it, which means, theoretically, the ratio between negative and positive examples is about $100:1$. Using the method introduced in Section~\ref{sec:neglabels}, we can obtain a selected set of negative examples $D_n$ derived from push notifications and another set of negative examples $D_c$ derived from dispatcher calls. The set $D_c$ is about the same size as the positive examples $D_p$, while $|D_n| : |D_p| \approx 11:1$. When training the neural network, we always use all the examples from $D_p$ and $D_c$. However, we randomly sample a subset of examples from $D_n$ at each episode of the training. By doing this, we ensure that the negative examples from $D_n$ do not dominate the training set, and at the same time the ``more certain'' negative examples from $D_c$ gets emphasized more than $D_n$. This whole procedure leads to an overall ratio between negative and positive samples around $3.5:1$ in each single batch.

\paragraph{Hyperparameters}
We ran a grid search over the hyperparameters of the ML model on an offline recommendation task. In the search process, we used the data from March 2018 to October 2019 (i.e. not including the data we test PROOF on), where the first $7/8$ of the selected data are used as the training set and the last $1/8$ are used as the validation set. We show the search result in Table~\ref{tab:hyperparameters}.

\begin{table}[]
\centering
\caption{Hyperparameters tuning}
\label{tab:hyperparameters}
\begin{tabular}{@{}lll@{}}
\toprule
Hyperparameter                                                           & Values Attempted            & Value Chosen \\ \midrule
Adam learning rate                                                       & $10^{-5}, 10^{-4}, 10^{-3}$ & $10^{-3}$    \\ \midrule
\begin{tabular}[c]{@{}l@{}}L2 regularization \\ coefficient\end{tabular} & $10^{-6}, 10^{-4}, 10^{-3}$ & $10^{-4}$    \\ \midrule
Batch size                                                               & 1024, 256                   & 256          \\ \bottomrule
\end{tabular}
\end{table}

\end{document}